\documentclass[10pt]{arxiv}

\usepackage{amsmath,amsfonts,bm}

\def\eqref#1{equation~\ref{#1}}

\def\1{\bm{1}}

\DeclareMathAlphabet{\mathsfit}{\encodingdefault}{\sfdefault}{m}{sl}
\SetMathAlphabet{\mathsfit}{bold}{\encodingdefault}{\sfdefault}{bx}{n}

\usepackage{amsmath,amssymb}
\usepackage{booktabs}
\usepackage{graphicx}
\usepackage{microtype}
\usepackage{multirow}
\usepackage{placeins}
\usepackage{float}
\usepackage{needspace}
\usepackage{xcolor}
\usepackage{tikz}
\usepackage{pgfplots}
\usepackage{hyperref}
\usepackage{url}
\usepackage{fvextra}
\usepackage{fontawesome5}
\usetikzlibrary{arrows.meta,positioning,fit}
\pgfplotsset{compat=1.18}

\definecolor{csiBlue}{HTML}{0072B2}
\definecolor{csiGreen}{HTML}{009E73}
\definecolor{csiMagenta}{HTML}{CC79A7}
\definecolor{csiOrange}{HTML}{E69F00}
\definecolor{csiSFT}{HTML}{D55E00}
\definecolor{csiGray}{HTML}{666666}
\definecolor{csiLightBlue}{HTML}{DCEEF8}
\definecolor{csiLightGreen}{HTML}{DDF2EA}
\definecolor{csiLightOrange}{HTML}{FBE9C8}

\newcommand{\bank}{\mathcal{B}}
\newcommand{\audit}{\mathcal{A}}
\newcommand{\update}{\mathcal{D}_{\mathrm{new}}}
\newcommand{\pp}{\text{pp}}
\newcommand{\learningrate}[1]{\mbox{\texttt{#1}}}

\title{Self-Preservation in Language Models}
\author[1]{Dongkyu Cho}
\author[1]{Rumi Chunara}
\author[2,3]{Ali Behrouz}
\affiliation[1]{New York University}
\affiliation[2]{Google Research}
\affiliation[3]{Cornell University}
\contribution{\email{dongkyu.cho@nyu.edu}\quad
\email{rumi.chunara@nyu.edu}\quad
\email{alibehrouz@google.com}}
\graphicspath{{./}}
\titlespacing*{\section}{0pt}{12pt plus 2pt minus 2pt}{6pt}
\titlespacing*{\subsection}{0pt}{10pt plus 2pt minus 2pt}{4pt}
\titlespacing*{\subsubsection}{0pt}{8pt plus 2pt minus 2pt}{3pt}
\hypersetup{pdftitle={Self-Preservation in Language Models},pdfauthor={Dongkyu Cho; Rumi Chunara; Ali Behrouz}}

\abstract{
Continual learning (CL) seeks to adapt pretrained models while preserving
prior capabilities. Yet the tasks supplied for preservation cover only part
of what a model knows, while forgetting can extend beyond them. Recent work
shows that language models can supplement this limited evidence by generating
useful replay without access to their original training data. This raises a question about self-preservation: can a model help choose preservation targets that protect its own capabilities more broadly? To study this question, we organize task descriptions and their examples into Task Banks, whose entries
can be selected and combined to guide replay generation by a frozen model.
Continual instruction tuning with Qwen models of varying sizes yields three
findings. First, the new task and learning rate shape which prior
capabilities degrade, while tracking previously correct answers reveals
losses that average accuracy can conceal. Second, generated replay reduces
forgetting on tasks outside the bank. Combining partial banks strengthens
this protection, improving mean retention of previously correct answers over
either bank alone across the tested sizes. Third, the model can also help
choose replay targets: selection informed by the new task improves mean
held-out retention over random selection at 4B, with similar new-task gains,
although smaller models show no corresponding advantage. Under a fixed
example budget, however, replacing new-task examples with replay improves
preservation at the cost of smaller new-task gains. Together, these findings
show that models can support broader self-preservation by generating replay
data and, in some settings, choosing which tasks to replay.

}
\begin{document}
\maketitle

\vspace{6pt}
\begingroup
\setlength{\parindent}{0pt}
\sffamily
\textcolor{themefg}{\faGithub\enspace\textbf{Code:}}\enspace
\href{https://github.com/umamicode/self-preservation}
{github.com/umamicode/self-preservation}\par
\endgroup

\section{Introduction}
\label{sec:introduction}

\begin{figure}[!t]
\centering
\includegraphics[width=\linewidth]{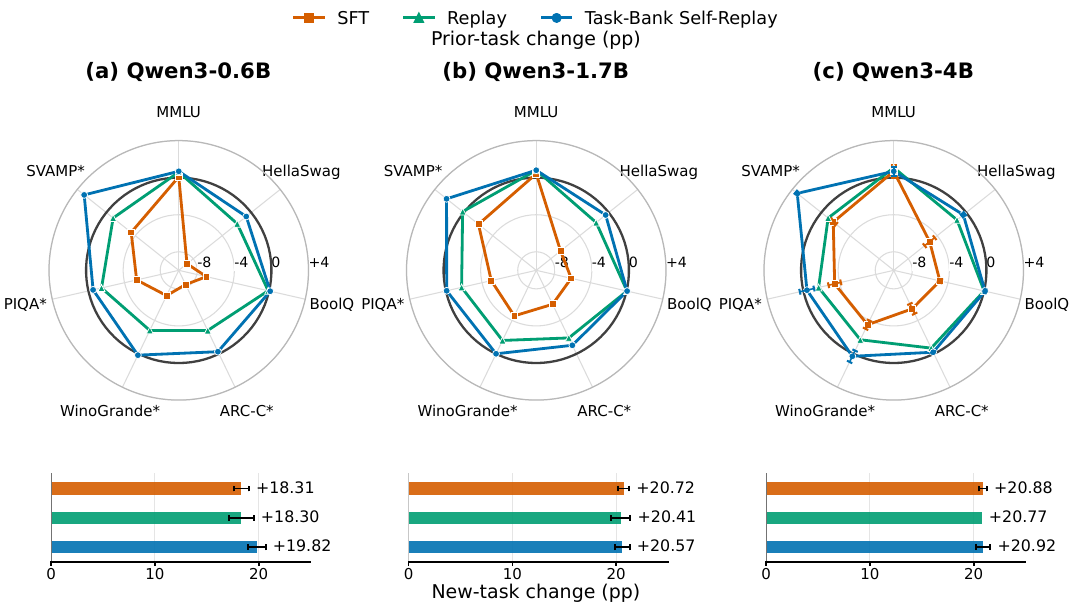}
\caption{\textbf{Generated replay across supplied targets can protect held-out capabilities while retaining new-task gains.}
Top: prior-task accuracy changes. Bottom: incoming-task gains for
Qwen3-0.6B, 1.7B, and 4B after one epoch on 1,024 MedQA examples at
$\learningrate{1e-5}$.
\textcolor{csiSFT}{SFT} uses incoming data only.
\textcolor{csiGreen}{Replay} uses real prior-task examples and labels.
\textcolor{csiBlue}{Task-Bank Self-Replay} uses model-generated questions and
answers. Both replay methods cover MMLU, GSM8K, HellaSwag, and BoolQ. The plot
shows seven prior tasks. Stars mark held-out tasks: ARC-C, WinoGrande, PIQA,
and SVAMP. Points are five-seed means. Displayed error bars are $\pm1$ SE.
The zero ring denotes pre-update accuracy, and all radial panels span
$-10$ to $+4$ pp.}
\label{fig:overview}
\end{figure}

Language models acquire broad capabilities through costly pretraining, yet
need further training to incorporate new information or specialize in new
domains. For these updates to produce cumulative improvement, models must
learn new behavior without losing useful capabilities they already possess.
Fine-tuning can instead degrade prior behavior, a phenomenon known as
catastrophic forgetting. Continual learning (CL) mitigates this problem
through replay and updates constrained by evidence of prior behavior
\citep{kirkpatrick2017overcoming,chaudhry2019tiny}. When the original training
data are unavailable, however, the tasks and examples supplied for
preservation cover only part of what the model should retain
\citep{cho2026continuallearningselfprobegradients}.
Figure~\ref{fig:overview} illustrates this mismatch: new-task gains coexist
with losses across prior tasks, and replay from real data reduces these
losses without eliminating them.

Recent work shows that language models can supplement limited preservation
evidence via self-generated data
\citep{sun2019lamol,huang2024ssr,marek2026selfreplay}.
These findings motivate the idea of \emph{self-preservation}: using a model's
own knowledge to help retain its capabilities during adaptation. Generating
more examples within a supplied task can strengthen the evidence for that
task, but does not by itself establish protection elsewhere. We therefore
examine whether combining preservation targets improves retention beyond
the supplied inventory, and whether information about the incoming task
helps a model select useful targets.

To study these choices, we organize supplied preservation targets into
\emph{Task Banks}, compact collections of task descriptions and optional
examples. Task-Bank Self-Replay uses a frozen pre-update model to generate
replay for selected entries and mixes these data with new-task examples
during fine-tuning. The bank separates the specification of what to replay
from the examples generated for it. Combining partial banks tests whether
broader supplied coverage improves preservation. Within a fixed bank, we
compare random selection with model selection based on either the bank
alone or the bank and information about the incoming task. These comparisons
test the model's contribution to choosing targets as well as generating
replay data.

Testing protection beyond the bank requires evaluation on behavior that did
not guide replay construction. We reserve held-out tasks whose examples,
labels, descriptions, and scores are excluded from bank construction and
target selection. We also examine individual tasks and previously correct
answers: gains on some tasks can offset losses on others, and newly
corrected answers can offset losses of previously correct answers
\citep{harmon2025mappingposttrainingforgettinglanguage}. Alongside task-level
accuracy changes, we report \emph{Retention}, the fraction of initially
correct answers that remain correct after training, where paired
predictions are available \citep{yang2026cpl}.
We assess preservation together with new-task gains. Our instruction-tuning experiments span Qwen models from 0.6B to 4B
\citep{yang2025qwen3} and yield three main findings:
\begin{itemize}
\setlength{\itemsep}{2pt}
\setlength{\parskip}{0pt}
\item \textbf{Training choices shape forgetting, and aggregate accuracy can conceal it.}
The incoming task and learning rate affect which prior capabilities degrade.
Tracking individual tasks and previously correct answers reveals losses
that favorable averages can hide.
\item \textbf{Generated replay supports preservation beyond its supplied targets.}
Task-Bank Self-Replay reduces held-out losses while retaining new-task gains.
Combining two partial banks improves mean held-out Retention over either
bank alone at all three tested Qwen3 sizes.
\item \textbf{The model can help choose useful replay targets.}
In the tested 4B setting, selection informed by the incoming task improves
mean held-out Retention by 2.75 percentage points over random selection,
with similar new-task gains. The same advantage does not appear at the
smaller tested sizes.
\end{itemize}


\Needspace{8\baselineskip}
\section{Motivation and Method}
\label{sec:method}

\subsection{Preservation Beyond Replay Targets}
\label{sec:broad_forgetting}

Adapting to one task can change performance on others
\citep{wang2023trace}. A medical question-answering update, for example, changes both
reasoning and commonsense performance in Figure~\ref{fig:overview}. Gains on
the incoming task do not tell us whether the model has retained its broader
capabilities~\citep{yang2026cpl}. We also evaluate prior tasks, including those absent from the
data used to protect the model.

Limited past data restrict the range of behavior represented in preservation
evidence. Replaying retained examples supplies a direct learning signal for
their tasks \citep{chaudhry2019tiny}, while protection of other behavior
depends on transfer.
Generating more examples for the same targets adds depth to this evidence~\citep{marek2026selfreplay},
but does not by itself establish broader coverage. The amount of rehearsal
and the choice of targets are therefore distinct design decisions.

Task Banks record which behaviors the replay is intended to rehearse,
without requiring a complete inventory of the model's capabilities. We use
them to test whether different targets provide complementary protection
and whether incoming-task information helps choose among them. We evaluate the bank-composition and target-selection comparisons on
held-out tasks excluded from replay construction and selection. Performance
on these tasks tests whether protection generalizes beyond the bank.

\subsection{Task Banks}
\label{sec:task_banks}

A Task Bank $\mathcal{B}=\{b_1,\ldots,b_m\}$ contains compact task entries:
a name, a description of the behavior and input/output format, and optional
demonstrations. Entries may
describe overlapping skills. Replay can use one entry, a selected subset
$S\subseteq\{1,\ldots,m\}$, or all supplied entries (the full bank). SFT uses incoming data only.
Single-Task Replay rehearses one supplied task. Replay uses stored examples
and labels. Task-Bank Self-Replay instead obtains replay answers from a frozen
pre-update model.

We select the entries to rehearse before generating replay. We compare random selection, selection using the bank alone, and
selection also informed by the incoming task. The selector chooses entries,
the frozen reference model generates replay, and the learner receives the
update. These roles remain distinct even when they use the same checkpoint.
Appendix~\ref{app:target_selection} specifies the selection inputs and prompts.

Task-Bank Self-Replay generates its rehearsal data before the learner update.
The endpoint, bank-composition, and target-selection studies generate both
questions and answers for supplied targets. The fixed-example-allocation
study instead generates answers to supplied questions, conditioned on two
verified demonstrations while withholding the question's source answer
(Appendix~\ref{app:self_replay}). The learner minimizes supervised fine-tuning
loss on incoming examples mixed with replay. The reference model stays
frozen. Appendix~\ref{app:replay_training} gives the mixture objective,
allocation weights, and study-specific controls.

Evaluation separates new-task learning from preservation on replay and
held-out tasks. Held-out examples, labels, descriptions, and scores are
excluded from bank construction and target selection. Exploration protocols instead ask the model to propose new task
descriptions, with or without access to the retained inventory, rather than
select from the supplied entries. Appendix~\ref{app:bank_details} describes the generation
interfaces, exploration procedures, and analysis-only comparison conditions.

\section{Experiments}
\label{sec:experiments}

We test how generated replay, bank composition, and target selection affect
preservation beyond supplied targets and new-task learning. \emph{Held-out
tasks} are excluded from replay construction and target selection, so their
results measure protection beyond the replay inventory.

\subsection{Experimental Setting}
\label{sec:setting}

\paragraph{Models and benchmarks.}
We use Qwen3-0.6B, Qwen3-1.7B, and Qwen3-4B \citep{yang2025qwen3},
with Qwen3.5-4B \citep{qwen2026qwen35} included in the forgetting diagnostics.
Incoming tasks are MedQA \citep{jin2021medqa} and C-STANCE from TRACE
\citep{wang2023trace}. The Task Bank contains MMLU, HellaSwag, GSM8K, and
BoolQ. ARC-Challenge (ARC-C), WinoGrande, PIQA, and SVAMP are held out from
replay construction and target selection. Appendix~\ref{app:protocol}
gives study-specific model assignments, task roles, and evaluation formats.

\paragraph{Training protocols.}
The endpoint, bank-composition, and target-selection studies use one epoch,
a learning rate of $\learningrate{1e-5}$, and five training seeds. Their
MedQA comparisons use 1,024 incoming examples. Endpoint settings are selected
during development and held fixed for evaluation, following
\citet{cha2025hyperparameterscontinuallearningreality}. A separate controlled
forgetting sweep varies the model, incoming task, and learning rate
$\{\learningrate{1e-7},\learningrate{1e-6},\learningrate{1e-5}\}$,
using one epoch of full-parameter AdamW. We also test a fixed example
allocation in which replay replaces incoming examples. That comparison
controls nominal rows and update counts, while token exposure and generation cost
can differ. Appendices~\ref{app:optimization} and~\ref{app:hyperparameters}
give optimizer settings, sweep grids, and study-specific allocations.

\paragraph{Evaluation.}
We report incoming-task gains and prior-task accuracy changes in percentage
points on the same pre/post evaluation examples. Where matched predictions
are available, we also report \emph{Retention} (correct $\rightarrow$
correct): the percentage of initially correct answers that remain correct
after the update, following \citet{cho2026continuallearningselfprobegradients}.
This separates preserved answers from newly corrected errors. Error bars
denote standard errors across training seeds. Appendix~\ref{app:measurements}
gives task aggregation and Retention eligibility rules.

\subsection{Experimental Results}
\label{sec:results}
\label{sec:forgetting}

\Needspace{5\baselineskip}
\subsubsection{Aggregate accuracy can conceal lost behavior}
\label{sec:broad_forgetting_results}
\label{sec:paired_answer_changes}

\begin{figure}[!htbp]
\centering
\includegraphics[width=0.65\linewidth]{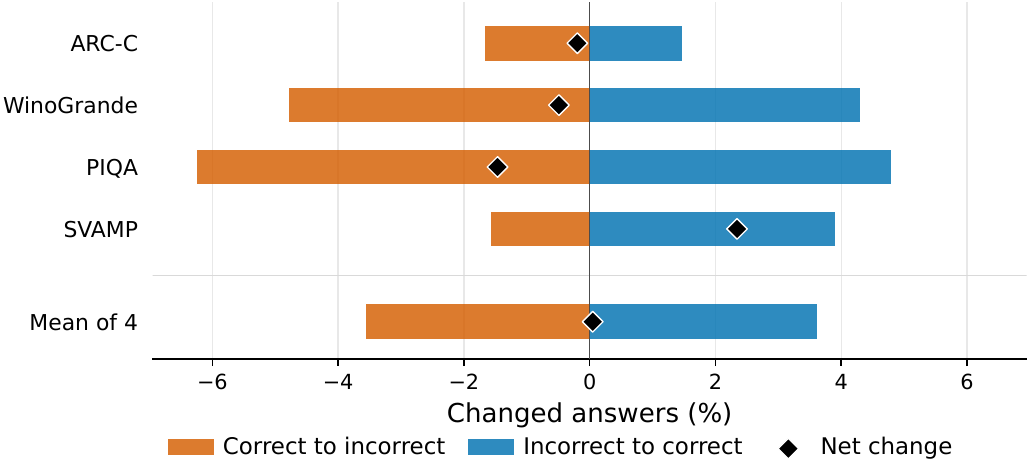}
\caption{\textbf{Nearly unchanged accuracy can conceal substantial loss of previously correct answers.}
Leftward bars show correct-to-incorrect transitions. Rightward bars show
incorrect-to-correct transitions. Diamonds show their net effect on accuracy.
The four-task mean loses 3.56\% and gains 3.61\% of examples, for a net
$+0.05$ pp change. Qwen3-4B learns MedQA using 256 input rows, one epoch,
$\learningrate{1e-6}$, and two seeds. Each task has 512 paired evaluation
items. Bars use all items as the denominator. Retention (correct
$\rightarrow$ correct) conditions on initially correct items.}
\label{fig:answer_turnover}
\end{figure}

\looseness=-1
Figure~\ref{fig:overview} shows incoming-task gains alongside prior-task
losses. Within a task, newly corrected answers can also offset the loss of
previously correct answers
\citep{harmon2025mappingposttrainingforgettinglanguage}.
Averaged over four held-out tasks and two seeds, 3.56\% of examples become
incorrect and 3.61\% become correct, leaving only a $+0.05$ pp net change
(Figure~\ref{fig:answer_turnover}). We therefore report paired Retention
alongside task-level accuracy changes where matched predictions are available.

\FloatBarrier
\Needspace{5\baselineskip}
\subsubsection{Broad replay improves preservation outside its inventory}
\label{sec:single_task}

\begin{figure}[!htbp]
\centering
\includegraphics[width=0.78\linewidth]{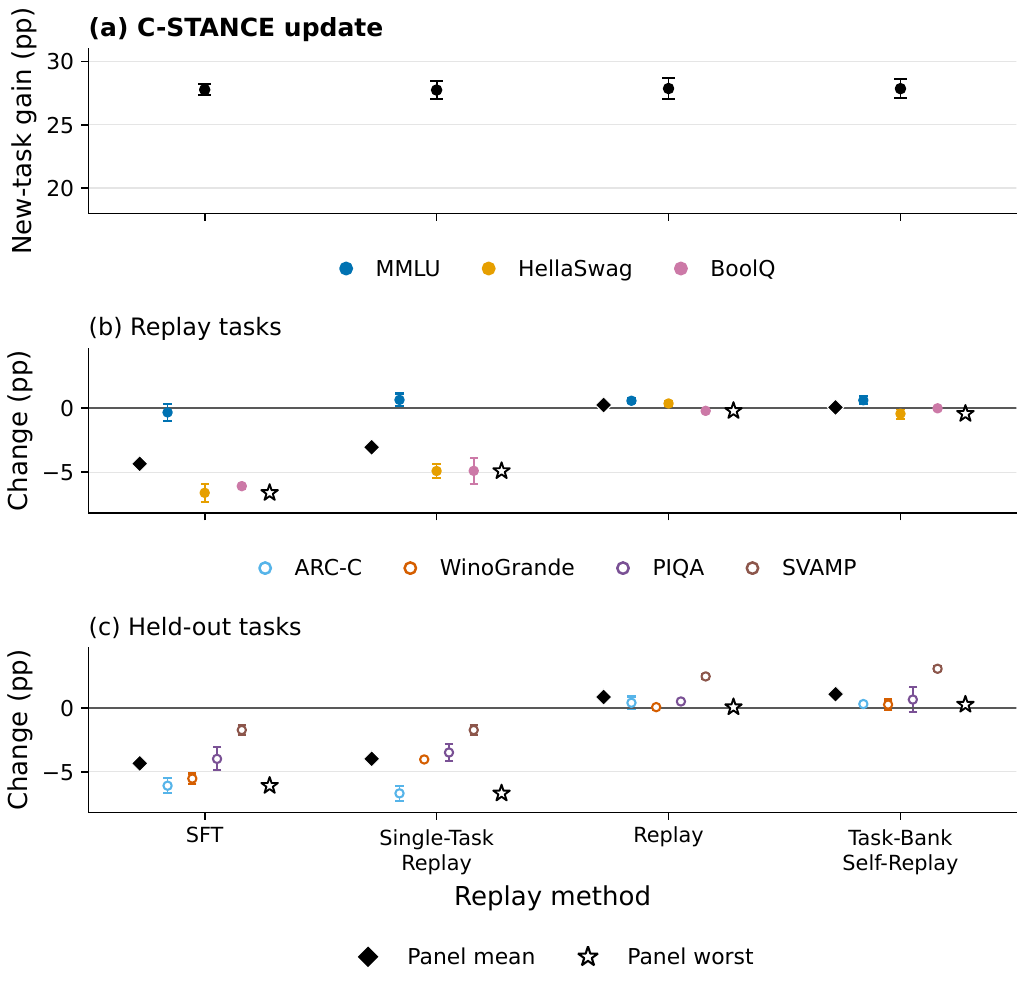}
\caption{\textbf{Broad replay protects held-out tasks better than single-target replay while retaining similar new-task gains.}
SFT uses incoming data only. Single-Task Replay rehearses one prior target.
Replay uses real examples and labels from MMLU, GSM8K, HellaSwag, and BoolQ.
Task-Bank
Self-Replay generates replay questions and answers for that bank.
Rows show incoming gain, changes on three displayed replay tasks, and changes
on four held-out tasks. Qwen3-4B learns C-STANCE for one epoch at
$\learningrate{1e-5}$. Circles are five-seed means with $\pm1$ SE relative to
fixed pre-update scores. Diamonds and stars summarize the mean and minimum
of the displayed task means. Zero denotes pre-update performance.}
\label{fig:single_task}
\end{figure}

\looseness=-1
Broad replay protects held-out tasks more effectively than replay restricted
to one target in the C-STANCE comparison (Figure~\ref{fig:single_task}).
Mean held-out accuracy changes are $-4.34\,\pp$ for SFT and $-3.99\,\pp$ for
Single-Task Replay, compared with $+0.86\,\pp$ for Replay and $+1.07\,\pp$
for Task-Bank Self-Replay. Incoming gains remain close, from $+27.74$ to
$+27.86\,\pp$.
Table~\ref{tab:manual_cstance} gives absolute scores. Figure~\ref{fig:overview}
compares three model sizes on MedQA.

\subsubsection{Combining supplied targets improves mean held-out Retention}
\label{sec:bank_composition}
\label{sec:topic_expansion_method}
\label{sec:topic_expansion_results}

\begin{figure}[!htbp]
\centering
\includegraphics[width=\linewidth]{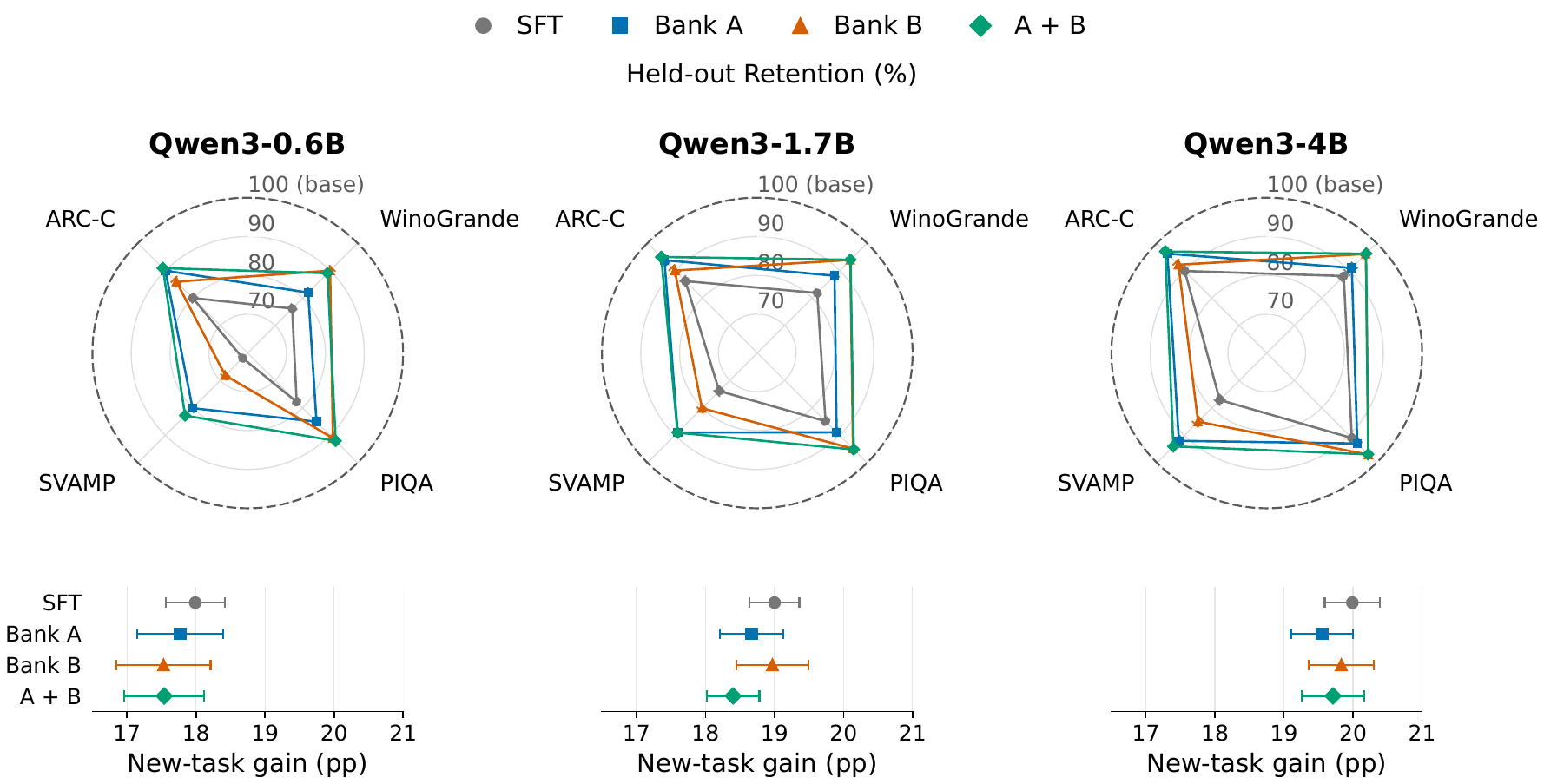}
\caption{\textbf{Combining partial banks improves mean held-out Retention, although the best bank varies by task.}
Bank A contains MMLU and GSM8K. Bank B contains HellaSwag and BoolQ. A+B
combines all four. Replay questions and answers are model-generated.
Top: Retention (correct $\rightarrow$ correct) on ARC-C, WinoGrande, PIQA,
and SVAMP, all held out from replay construction and target selection.
The dashed outer ring is the frozen model's 100\% Retention reference,
not 100\% accuracy. The radial range is 60--100\%.
Bottom: MedQA accuracy gain. Each model uses the Figure~\ref{fig:overview}
setting: 1,024 incoming examples, one epoch, and $\learningrate{1e-5}$.
Points and bars are means $\pm1$ SE across five seeds.}
\label{fig:bank_width_heldout}
\end{figure}

Combining the two supplied partial banks broadens the replay inventory and improves
mean held-out Retention in all three tested models (Figure~\ref{fig:bank_width_heldout}).
The combined bank reaches $88.70\%$, $93.28\%$, and $96.01\%$ Retention,
exceeding the better partial bank by 4.42, 3.32, and 2.99 pp at 0.6B, 1.7B,
and 4B. Incoming gains are 0.45, 0.60, and 0.28 pp below SFT. The best bank
varies by held-out task, while the combined bank has the best mean. Combining
supplied targets therefore improves protection beyond the bank without
requiring the model to discover new tasks.

\begin{figure}[!htbp]
\centering
\includegraphics[width=\linewidth]{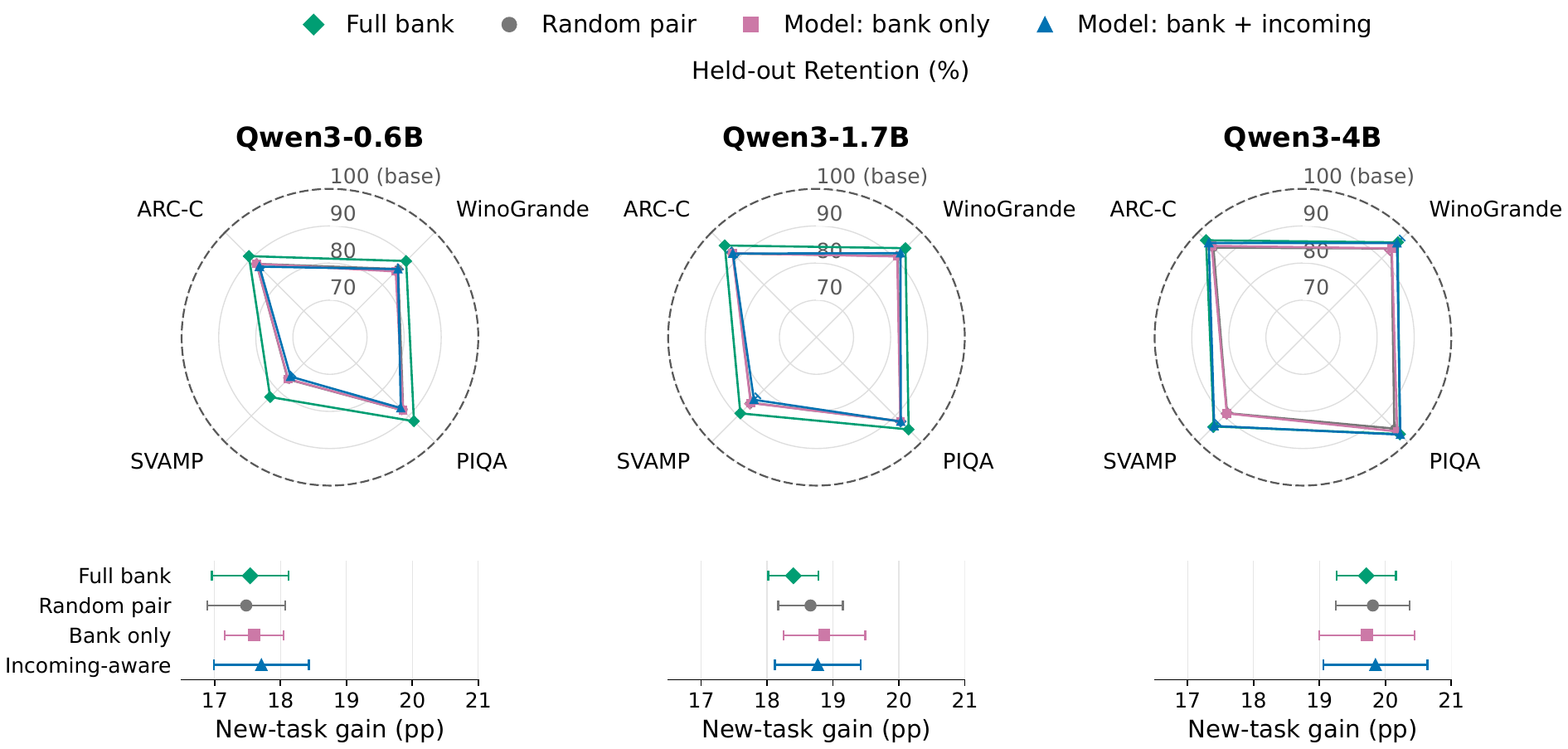}
\caption{\textbf{Incoming-aware selection improves held-out Retention at 4B and approaches the full-bank reference with two targets.}
Random, bank-only, and incoming-aware selection each choose two entries from
MMLU, GSM8K, HellaSwag, and BoolQ before replay generation. Bank-only selection
receives the inventory. Incoming-aware selection also receives incoming-task
information. The full bank uses all four entries. No corresponding mean
advantage over random selection appears at 0.6B or 1.7B.
Top: held-out Retention (correct $\rightarrow$ correct), with a 60--100\%
radial range and a dashed 100\% pre-update reference. Bottom: MedQA accuracy
gain. Training follows Figure~\ref{fig:overview}: 1,024 incoming examples,
one epoch, and $\learningrate{1e-5}$. Points and bars are five-seed means
$\pm1$ SE. All four evaluation tasks are excluded from selection and replay.}
\label{fig:bank_selection_heldout}
\end{figure}

\subsubsection{Incoming-task information can improve target selection}
\label{sec:target_selection}

When replay uses only two bank entries, incoming-task information improves
target selection at 4B, but gives no corresponding advantage at the smaller
tested sizes
(Figure~\ref{fig:bank_selection_heldout}). At 4B, incoming-aware selection
reaches $95.69\%$ mean held-out Retention, compared with $92.94\%$ for random
selection and $93.42\%$ for bank-only selection. Its incoming gain is 0.04 pp
above random selection, and Retention is within 0.32 pp of the full bank
while using two of its four targets. The Retention difference from random
selection is $-0.76\,\pp$ at 0.6B and approximately zero at 1.7B. The observed
selection advantage is limited to the tested 4B configuration. Learner,
selector, and replay-generator strength are not varied independently.
Appendix~\ref{app:target_selection} describes the selection interfaces.

Proposed additions to the inventory are examined separately in the
bank-aware discovery pilot (Appendix~\ref{app:bankaware_negative_pilot}).
This single-arm pilot tests one operating point and does not estimate the
incremental preservation value of discovery relative to a fixed bank.

\Needspace{5\baselineskip}
\subsubsection{Preservation under a fixed example allocation}
\label{sec:protection}

\begin{figure}[!htbp]
\centering
\includegraphics[width=\linewidth]{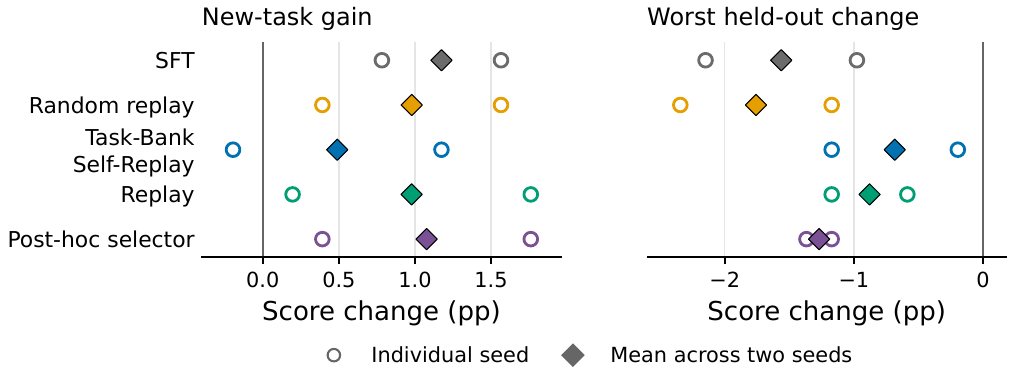}
\caption{\textbf{Replay can reduce the worst held-out loss while also reducing new-task gains under a fixed example allocation.}
Left: incoming accuracy gain. Right: mean seed-wise worst held-out accuracy
change. Higher is better. Qwen3-4B learns MedQA at $\learningrate{1e-6}$ for
one epoch and 16 updates. SFT uses 256 incoming rows. Replay arms use 197
incoming and 59 replay rows. Task-Bank Self-Replay generates answers to
supplied questions. Replay uses stored examples and labels. Random replay uses one
bank task. The post-hoc selector uses observed SFT damage as an analysis
reference. Open circles are seeds. Diamonds average two seeds. Input rows
and updates are matched, but tokens and generation cost are not.
Table~\ref{tab:protection} gives \mbox{paired forgetting and complete scores.}}
\label{fig:meta_bank}
\end{figure}

Replacing incoming examples with replay exposes a preservation--acquisition
trade-off (Figure~\ref{fig:meta_bank}). With 59 of 256 input rows allocated to
replay, Task-Bank Self-Replay improves the mean seed-wise worst held-out
change from $-1.56$ to $-0.68\,\pp$, while MedQA gain falls from $+1.17$ to
$+0.49\,\pp$. Replay gives an intermediate worst held-out change of
$-0.88\,\pp$ and an incoming gain of $+0.98\,\pp$. The preservation
benefit cannot be attributed to replay content alone, since replay also
reduces incoming exposure.

The methods share nominal input allocation and optimizer steps, while token
exposure and generation cost can differ. This study uses the
supplied-question generation protocol in Appendix~\ref{app:self_replay}.

Table~\ref{tab:protection} shows that forgetting among initially correct
answers can decrease even when the worst-task accuracy change worsens.
This fixed-allocation comparison does not establish a better
acquisition--retention frontier.

\FloatBarrier
\subsubsection{The incoming task and learning rate shape preservation}
\label{sec:adaptation_conditions}

\begin{figure}[!htbp]
\centering
\includegraphics[width=\linewidth]{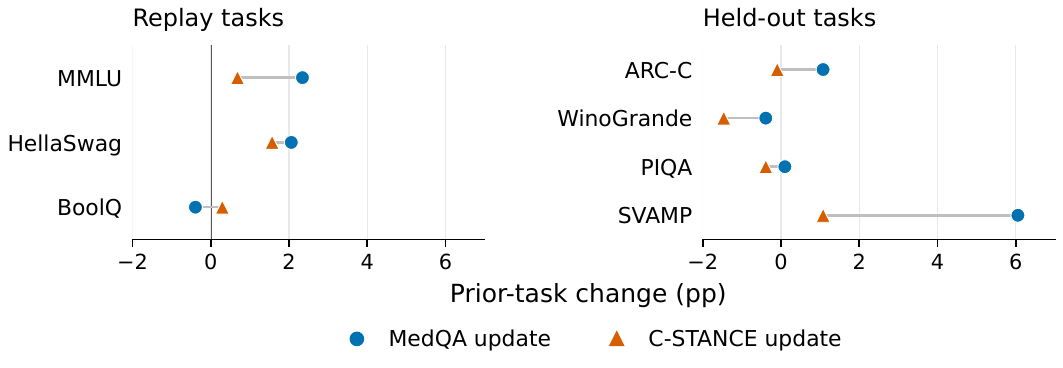}
\caption{\textbf{Different incoming tasks cause different patterns of forgetting under the same training protocol.} Qwen3.5-4B is updated with SFT on MedQA or C-STANCE at $\learningrate{1e-6}$,
holding the model and optimization protocol fixed. Paired markers show each
capability's score change averaged across two seeds. Connecting segments
compare update conditions, not training checkpoints. The two panels group
three displayed bank tasks and four held-out tasks on the same horizontal
scale. These groups identify evaluation tasks. Neither supplies replay in
these SFT runs. The zero line denotes the pre-update model. The respective
new-task gains are $+4.7\,\pp$ for MedQA and $+7.8\,\pp$ for C-STANCE.}
\label{fig:update_profiles}
\end{figure}

\looseness=-1
\paragraph{Incoming-task effects.}
The incoming task changes the prior-behavior profile even when the model and
optimization protocol are held fixed. At $\learningrate{1e-6}$,
Qwen3.5-4B gains $+4.7\,\pp$ on MedQA and $+7.8\,\pp$ on C-STANCE, but the
changes across prior tasks differ (Figure~\ref{fig:update_profiles}).
SVAMP accuracy increases by about six points after the MedQA update and
about one after the C-STANCE update. The effect is heterogeneous across the
evaluated capabilities, \mbox{rather than a uniform shift.}

\begin{figure}[!htbp]
\centering
\includegraphics[width=\linewidth,height=0.82\textheight,keepaspectratio]{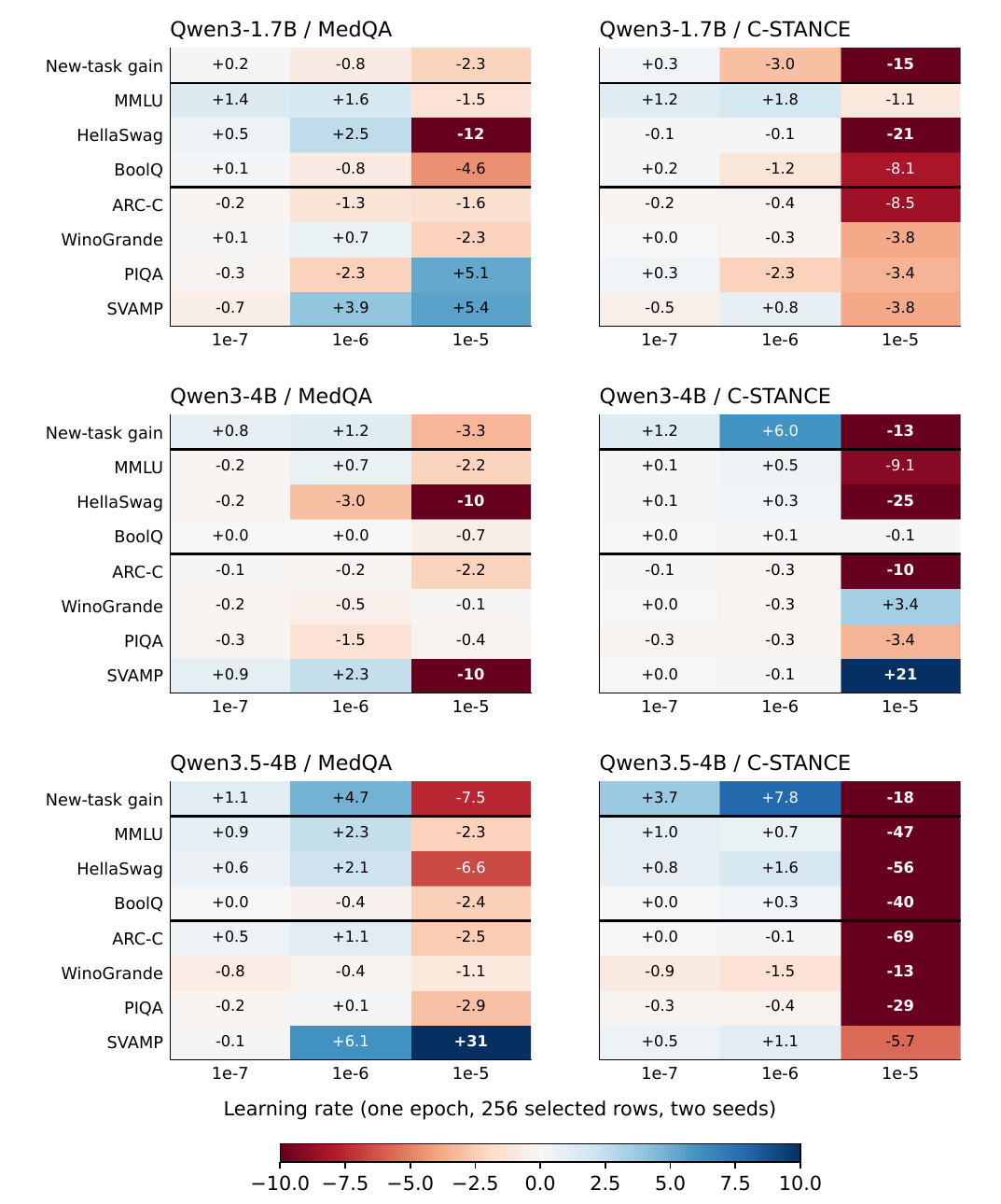}
\caption{\textbf{A learning rate that supports useful new-task learning can still cause uneven losses across prior capabilities.}
The complete main sweep contains 36 SFT runs: three model rows, two update-task
columns, three learning rates, and two seeds. Each run uses 256 selected input
rows for one epoch. Cells show seed-mean percentage-point score change from
the pre-update model. The top row of each panel is new-task gain. Horizontal
rules separate it from three displayed bank tasks and four held-out tasks.
Color uses one common scale saturated at
$\pm10$ pp. Printed values preserve larger magnitudes. At $\learningrate{1e-5}$, several
conditions lose new-task accuracy as well as prior-task accuracy, indicating
failed adaptation rather than a necessary retention--acquisition trade-off.
Bank tasks are replay-eligible evaluation tasks. No replay is used in these SFT runs.
These are endpoint comparisons across learning rates, not training trajectories.
The full capability profiles appear in Appendix~\ref{app:lr_profiles}.}
\label{fig:lr_example}
\end{figure}

\looseness=-1
\Needspace{10\baselineskip}
\paragraph{Learning-rate effects.}
The complete 36-run sweep shows that learning rate affects both
incoming-task learning and prior-task changes
(Figure~\ref{fig:lr_example}). Optimization is a known determinant of
forgetting \citep{marek2026selfreplay}. Here we examine the profiles under
short, fixed-exposure updates. Under this 256-row, one-epoch protocol, the
$\learningrate{1e-5}$ conditions reduce incoming-task accuracy across all six
model--task combinations, indicating failed adaptation.
Lower-rate conditions include useful incoming-task gains accompanied by
selective prior-task losses. Evaluating both outcomes distinguishes
preservation during learning from \mbox{small changes caused by limited adaptation.}

Lower learning rates can improve both new-task learning and preservation
under the same update budget. The separate 1,024-example Qwen3-1.7B C-STANCE
sweep in Table~\ref{tab:cstance_1p7b_lr} achieves $+12.01\,\pp$ incoming gain
and $97.02\%$ mean held-out Retention at $\learningrate{1e-7}$.
At $\learningrate{3e-6}$, these fall to $+9.67\,\pp$ and $82.22\%$.
This five-rate study uses 64 updates and is distinct from the three-rate,
256-row heatmap. Appendix~\ref{app:hyperparameters} describes the additional
0.6B allocations. Appendix~\ref{app:lr_profiles} provides full profiles.

\begin{table}[!htbp]
\centering
\small
\setlength{\tabcolsep}{5pt}
\begin{tabular}{lrrrr}
\toprule
Learning rate & \shortstack{Incoming test\\$\Delta$ accuracy}
& \shortstack{Held-out mean\\$\Delta$ accuracy}
& \shortstack{Worst held-out\\$\Delta$ accuracy}
& \shortstack{Held-out\\Retention (\%)} \\
\midrule
$\learningrate{1e-7}$ & $+12.01 \pm 0.10$ & $-0.10 \pm 0.10$ & $-0.49 \pm 0.10$ & $97.02 \pm 0.05$ \\
$\learningrate{3e-7}$ & $+3.32 \pm 5.08$ & $-0.73 \pm 0.34$ & $-2.15 \pm 0.39$ & $94.65 \pm 0.87$ \\
$\learningrate{1e-6}$ & $+10.45 \pm 1.86$ & $-1.98 \pm 0.46$ & $-9.57 \pm 0.59$ & $89.60 \pm 0.38$ \\
$\learningrate{3e-6}$ & $+9.67 \pm 0.49$ & $-3.03 \pm 0.24$ & $-19.63 \pm 0.88$ & $82.22 \pm 0.25$ \\
$\learningrate{1e-5}$ & $-2.05 \pm 0.88$ & $-28.00 \pm 3.00$ & $-63.18 \pm 6.35$ & $32.88 \pm 4.56$ \\
\bottomrule
\end{tabular}
\caption{\textbf{A lower learning rate can improve both new-task learning and retention of prior correct answers.}
Qwen3-1.7B learns C-STANCE from 1,024 examples for one epoch (64 updates),
separate from the 256-row study in Figure~\ref{fig:lr_example}.
Values are means $\pm$ SE across seeds 0 and 1 at the FP32 final checkpoint.
Accuracy changes compare matched pre/post predictions and use percentage
points. ARC-C, WinoGrande, PIQA, and SVAMP each have 512 evaluation items.
Incoming test accuracy uses a separate 512-item panel. Worst change is the
minimum across held-out tasks within each seed, then averaged. Retention
(correct $\rightarrow$ correct) is averaged equally across held-out tasks.}
\label{tab:cstance_1p7b_lr}
\end{table}

\FloatBarrier
\section{Related Work}
\label{sec:related}

\paragraph{Forgetting and continual learning.}
Replay and interference-aware methods mitigate forgetting by rehearsing prior
experience or controlling updates that would disrupt it
\citep{sun2019lamol,riemer2019mer,aljundi2019mir}. MIR selects replay examples
according to anticipated interference \citep{aljundi2019mir}. Our selection
study instead chooses task entries before generating replay. With abundant
replay memory, the remaining challenges can shift toward plasticity and
computational cost \citep{cho2026forgetforgettingcontinuallearning}. Evaluation beyond the incoming task reveals changes that aggregate
performance can conceal. TRACE studies continual language-model adaptation
across diverse tasks and evaluates general abilities and instruction
following \citep{wang2023trace}. Large-scale post-training analyses separate
losses of previously correct behavior from gains that offset them in
aggregate performance
\citep{harmon2025mappingposttrainingforgettinglanguage}. We use these
task-level and paired views to evaluate preservation beyond a partial
inventory of targets.

\paragraph{Generated preservation evidence and task exploration.}
Models can supply preservation evidence when prior data are limited.
LAMOL rehearses previous tasks with generated pseudo-examples
\citep{sun2019lamol}. Self-Synthesized Rehearsal uses in-context demonstrations
to synthesize instances, refines their outputs, and selects diverse rehearsal
data \citep{huang2024ssr}. The Self-Replay study of
\citet{marek2026selfreplay} examines self-generated replay and its interaction
with capacity and optimization. CPLUS uses anchor and self-probe gradients
to regulate updates and demonstrates preservation beyond the supplied anchor
task \citep{cho2026continuallearningselfprobegradients}. Our focus is how the
organization of supplied targets affects the protection from generated replay. Generating candidate tasks raises a separate question about what to preserve.
Self-Instruct generates instructions and input--output examples for
instruction tuning \citep{wang2023selfinstruct}, while Automated Capability
Discovery proposes tasks to probe a subject model's capabilities
\citep{lu2025acd}. Task Banks provide an explicit inventory for studying the
preservation value of task composition and selection. Our main comparisons
use supplied targets. Model-proposed additions remain exploratory.
Appendix~\ref{app:related} extends these comparisons to task-free learning,
on-policy supervision, and capability self-knowledge.

\section{Discussion}
\label{sec:discussion}

\paragraph{Toward sustained self-preservation.}
Our experiments leave open how to maintain a useful inventory across
successive updates.
Proposed entries need demonstrated competence, distinct behavioral coverage,
and incremental preservation value. Answering a task correctly does not establish that the model can predict
update damage or choose effective replay. The single-arm
expansion pilot produces negative accuracy changes. The two-update,
one-seed pilot has mixed accuracy changes under a prescribed task addition
(Appendices~\ref{app:bankaware_negative_pilot} and~\ref{app:self_improvement}).
These findings motivate studying inventory revision alongside retention of
capabilities learned across successive updates.

\paragraph{Preservation under SFT.}
We study SFT because it is a standard method for adapting pretrained models
\citep{chen2024spin} and a demanding setting for preservation. Comparisons with on-policy
reinforcement learning and on-policy distillation report greater forgetting
under SFT in the evaluated settings
\citep{chen2026retaining,shenfeld2026sdft}. Holding the update rule fixed
lets us examine how replay source and target choice affect retention.
On-policy training can reduce forgetting, but some degradation can remain
\citep{shenfeld2026sdft}. Its different learning signals may also change
which preservation targets are useful. The same evaluation of held-out
retention and new-task learning therefore remains relevant across training
regimes. Further discussion regarding target choice, scope, and cost analysis appears in Appendix~\ref{app:limitations}.

\section{Conclusion}
\label{sec:conclusion}

We study self-preservation of language models: how they can use their own
knowledge to retain prior capabilities while learning new tasks. Task Banks
provide a simple way to examine preservation from an incomplete inventory
of targets. Generated replay reduces forgetting beyond the bank, and
combining partial banks improves mean held-out Retention. Selection informed
by the incoming task further helps in the tested 4B setting, with similar
new-task gains, but shows no corresponding advantage at smaller sizes.
These benefits require evaluation of both retained behavior and new-task
learning: average accuracy can conceal lost correct answers, and replay can
reduce new-task gains when it replaces incoming examples. Language models
can thus support broader preservation through generated data and, in some
settings, useful choices about what to replay.

\section*{Acknowledgments}
We thank NYU HPC for their generous support and assistance with computational
simulations and experiments. We also acknowledge support from the 2026
Google Research Awards, a TPU compute grant from Google's
\href{https://sites.research.google/trc/}{TPU Research Cloud (TRC) program},
and the TPU Builder's Program.

\small
\setlength{\bibsep}{0pt plus 0.5pt}
\bibliography{references}
\bibliographystyle{references}

\clearpage
\appendix
\normalsize

\section{Experimental Protocols and Supplementary Measurements}
\label{app:protocol}

\subsection{Capability changes and loss of previously correct answers}
\label{app:measurements}

Capability changes compare post-update accuracy with the pre-update model on
the same evaluation examples. We denote incoming-task gain by
$\Delta_{\mathrm{new}}$ and prior-task change by $\Delta_j$, both in percentage
points. We report individual tasks and equally weighted task groups, with
standard errors across training seeds where available.

\begin{figure}[!htbp]
\centering
\includegraphics[width=\linewidth]{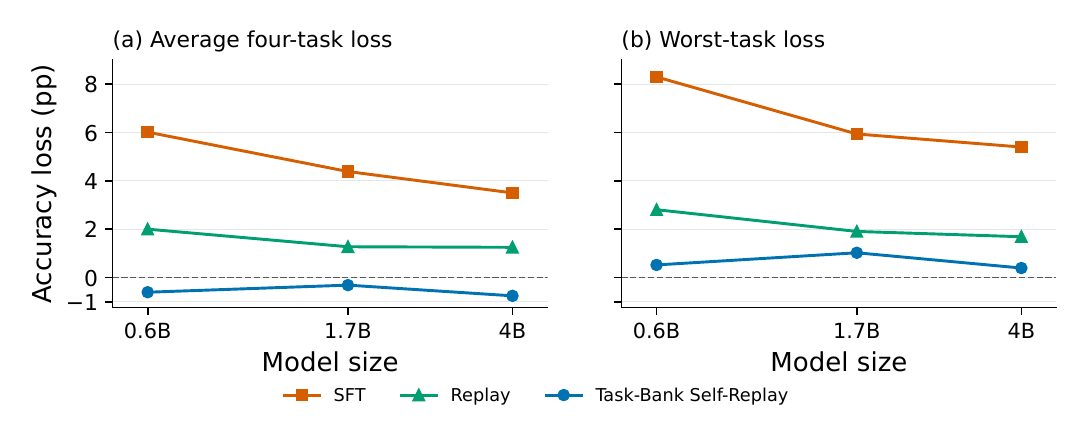}
\caption{\textbf{Broad replay reduces held-out accuracy losses across the tested sizes, but an average accuracy gain can still conceal an individual task loss.}
Qwen3 models are adapted to 1,024 MedQA examples for one epoch at learning
rate $\learningrate{1e-5}$. Average and worst-task accuracy loss summarize
ARC-C, WinoGrande, PIQA, and SVAMP. Each point summarizes five-seed task means.
Negative values indicate improvement. Lines connect measured sizes.}
\label{fig:scale_heldout_loss}
\end{figure}

We distinguish the minimum of capability means
from the average of seed-wise minima: the former summarizes the average
profile, while the latter asks how bad the worst capability is within each
run. Table~\ref{tab:cstance_1p7b_lr}, Figure~\ref{fig:meta_bank}, and
Table~\ref{tab:protection} average the worst held-out change within each seed.
The five-seed endpoint profiles in Figures~\ref{fig:single_task} and
\ref{fig:scale_heldout_loss} instead take the minimum of task means.
Paired seed-level uncertainty cannot be reconstructed \mbox{from the supplied endpoint aggregates.}

Forgetting among initially correct examples measures a different event: losing an answer
that the pre-update model previously got right. Let $\theta_0$ and $\theta_1$
denote the pre-update and post-update model parameters. For evaluation task $j$,
\begin{equation}
  F_j=\frac{|\{i:f_{\theta_0}(x_i)=y_i,\ f_{\theta_1}(x_i)\ne y_i\}|}
  {|\{i:f_{\theta_0}(x_i)=y_i\}|}.
  \label{eq:forgetting}
\end{equation}
Retention (correct $\rightarrow$ correct) is the complement of $F_j$,
expressed as a percentage. This quantity cannot be recovered from pre/post accuracy alone because newly
correct answers may offset previously correct answers that are lost.

The same paired predictions explain how gains and losses can cancel in net
accuracy, following the sample-wise view of
\citet{harmon2025mappingposttrainingforgettinglanguage}.
If $G_j$ is the number of previously incorrect answers that become correct,
$L_j$ the number of previously correct answers lost, and $N_j$ the evaluation
size, then $\Delta_j=100(G_j-L_j)/N_j$.
Figure~\ref{fig:answer_turnover} plots $100G_j/N_j$ and $-100L_j/N_j$ from
the recorded paired counts, using the same four held-out tasks for every summary.
We report raw behavioral transitions, not the chance-adjusted quantities
also developed by \citet{harmon2025mappingposttrainingforgettinglanguage}.
They do not by themselves establish erasure of underlying knowledge.

Conditioning on initially correct answers gives $F_j$ a different interpretation:
the fraction of previously correct behavior lost, rather than the fraction
of the full evaluation set that changes from correct to incorrect.
The fixed-example-allocation comparison in
Table~\ref{tab:protection} averages $F_j$ over the eligible ARC-C,
WinoGrande, and PIQA tasks. SVAMP does not meet the minimum number of initially
correct examples. This eligibility rule is specific to that study and is not
applied to the bank-composition or target-selection summaries.

\subsection{Capability inventory and access boundary}

The task split separates information used to update or protect the model from
information used only to test transfer. Table~\ref{tab:task_inventory} lists
the role and evaluation format of every Qwen task. Multiple-choice tasks are
scored by candidate-sequence loss, while GSM8K and SVAMP use short generation
with normalized numeric exact match. The main diagnostic profiles display
MMLU, HellaSwag, and BoolQ alongside the four held-out tasks.
Each replay task has distinct replay and
evaluation files, while held-out tasks have evaluation files only. The training and
replay-selection interfaces exclude held-out examples and scores.
Hyperparameter-selection procedures are distinguished by study in
Appendix~\ref{app:hyperparameters}.

\begin{table}[ht]
\caption{\textbf{Evaluation beyond replay targets tests the breadth of preservation.} Replay data is available only for
the replay tasks. The two update tasks supply training and validation data,
while the held-out tasks are evaluated \mbox{only after the update rule is frozen.}}
\label{tab:task_inventory}
\centering
\small
\setlength{\tabcolsep}{6pt}
\begin{tabular}{llll}
\toprule
Role & Task & Evaluation format & Replay data \\
\midrule
Update & MedQA & multiple choice & update data only \\
Update & C-STANCE & three-way choice & update data only \\
\midrule
Replay tasks & MMLU & multiple choice & yes \\
Replay tasks & HellaSwag & multiple choice & yes \\
Replay tasks & GSM8K & numeric generation & yes \\
Replay tasks & BoolQ & binary choice & yes \\
\midrule
Held-out evaluation & ARC-C & multiple choice & no \\
Held-out evaluation & WinoGrande & binary choice & no \\
Held-out evaluation & PIQA & binary choice & no \\
Held-out evaluation & SVAMP & numeric generation & no \\
\bottomrule
\end{tabular}
\end{table}

In the controlled studies, every capability uses a fixed pre/post evaluation set capped at 512 examples,
with at least 400 examples required after filtering. Generation is capped at
64 new tokens. In the fixed-example-allocation study, forgetting among initially correct
examples is aggregated only for tasks on which the pre-update model answers
at least 200 evaluation examples correctly. Other studies use their
respective evaluation panels and aggregation rules.

Example-bootstrap intervals quantify within-run evaluation variation, not
between-seed uncertainty. We use 1,000 paired resamples
with task-specific seeds formed by adding the zero-based evaluation task
index to 20260910.

\subsection{Model and optimization details}
\label{app:optimization}

The controlled studies fix checkpoint identity and check implementation
consistency before comparing updates. Table~\ref{tab:model_revisions} records
the pinned Qwen revisions. Qwen3.5-4B additionally required logit parity with
the official Transformers implementation before its sweep was admitted.
Main-sweep and protection results are checked for data integrity, task
coverage, and held-out exclusion after training and evaluation finish.

\begin{table}[ht]
\caption{\textbf{Fixed model revisions help make preservation comparisons interpretable.} Tensor parallelism (TP) and
TPU shape describe the main sweep and additional 0.6B sweep. The Qwen3-4B protection replication used
TP${=}4$ on v6e-4 \mbox{with the same pinned weights.}}
\label{tab:model_revisions}
\centering
\small
\setlength{\tabcolsep}{5pt}
\begin{tabular}{llll}
\toprule
Model & Revision & TP & TPU \\
\midrule
Qwen3-0.6B & \texttt{c1899de289a0} & 4 & v6e-4 \\
Qwen3-1.7B & \texttt{70d244cc86cc} & 4 & v6e-4 \\
Qwen3-4B & \texttt{1cfa9a720891} & 8 & v6e-8 \\
Qwen3.5-4B & \texttt{851bf6e806ef} & 8 & v6e-8 \\
\bottomrule
\end{tabular}
\end{table}

The checksum-validated main sweep and protection conditions use full-parameter AdamW with
$(\beta_1,\beta_2)=(0.9,0.95)$, $\epsilon=\learningrate{1e-8}$, zero weight decay, a
$5\%$ warmup when the update count permits it, and global-gradient clipping at
1.0. The micro-batch size is one with 16-step accumulation, maximum sequence
length 512, bfloat16 checkpoint storage, and model thinking disabled. Each
main sweep run selects 256 input rows for one epoch.

The protection comparison fixes nominal input allocation and optimizer steps,
not realized token exposure. Each run selects 256 input rows. In replay arms,
59 replay rows replace update rows rather than
adding optimization steps. In the ten protection-study runs, tokenization
retains 251--256 rows, including 54--59 replay rows in replay arms, and all runs use
16 optimizer steps. Realized row counts and token exposure therefore vary
across runs despite the shared nominal allocation.

\subsection{Hyperparameter sweeps and condition selection}
\label{app:hyperparameters}

The main sweep varies learning rate while holding the remaining configuration
fixed across conditions and seeds. It crosses Qwen3-1.7B, Qwen3-4B, and
Qwen3.5-4B with MedQA and C-STANCE, learning
rates $\{\learningrate{1e-7},\learningrate{1e-6},\learningrate{1e-5}\}$, and seeds 0 and 1. All 36 runs are
reported, rather than selecting a rate with the best prior-task accuracy.
The smaller-model follow-up additionally evaluates $\learningrate{3e-7}$ and
$\learningrate{3e-6}$ to locate a regime with useful new-task gains.

The comparison at similar new-task gains selects operating points by these gains,
not by prior-task accuracy. Figure~\ref{fig:profiles} includes
conditions in the $+4$ to $+10\pp$ C-STANCE new-task gain window, preferring
more completed seeds, then a gain closer to $+7\pp$, and finally a lower
learning rate. This plotting rule is descriptive and does not provide an
independent test of a tuned algorithm. The full measured profiles remain
available in Appendix~\ref{app:lr_profiles}.

The fixed-example-allocation replay study freezes one selected condition across
methods and seeds. It uses Qwen3-4B on MedQA at $\learningrate{1e-6}$, with
two seeds and 16 optimizer updates, and reuses the corresponding SFT main
sweep runs. Replay arms allocate 197 incoming and 59 replay rows before
tokenization. This condition was promoted after the SFT screen. It is not a
separately held-out development scenario.

\looseness=-1
The five-seed endpoint study uses a separate protocol with two phases:
hyperparameter tuning and evaluation. Following
\citet{cha2025hyperparameterscontinuallearningreality}, we selected
$\learningrate{1e-5}$ during tuning and held this learning \mbox{rate fixed during evaluation.}

The bank-composition and target-selection studies use the nominal MedQA
setting of Figure~\ref{fig:overview}: Qwen3-0.6B, Qwen3-1.7B, and Qwen3-4B,
1,024 incoming examples, one epoch at $\learningrate{1e-5}$, and five training
seeds. These studies report paired Retention separately from the endpoint
accuracy summaries.

\paragraph{Additional 0.6B allocations and evaluation panels.}
The additional Qwen3-0.6B C-STANCE calibration shown in
Figures~\ref{fig:profiles} and~\ref{fig:lr_atlas} uses 256 allocated rows.
Its $\learningrate{3e-6}$ condition averages two seeds with an incoming gain
of $+7.1\,\pp$. It is separate from the larger-allocation follow-ups below.

A set of 17 Qwen3-0.6B follow-up runs uses 1,024 examples and 64 updates:
seven MedQA runs on TPU and ten C-STANCE runs on an H100. At
$\learningrate{3e-6}$, the two reported MedQA seeds improve incoming
development accuracy by $+7.42$ and $+8.59\,\pp$ while changing held-out mean
accuracy by $-3.81$ and $-5.57\,\pp$. The two C-STANCE seeds improve
development accuracy by $+5.47$ and $+9.38\,\pp$ while changing held-out mean
accuracy by $-2.69$ and $-3.56\,\pp$. Their separate incoming test gains are
$+11.33$ and $+11.91\,\pp$. These comparisons use their respective matched
baselines and are analyzed separately from the calibration and main sweeps.

\subsection{Five-seed single-task endpoint study}
\label{app:manual_scores}

The endpoint study evaluates Qwen3-4B after one epoch at learning rate
$\learningrate{1e-5}$, with means and standard errors across five seeds. It contains two
update tasks and four methods per task. Table~\ref{tab:manual_cstance}
shows the C-STANCE absolute endpoint scores, with the pre-update model
score shown separately. This study reports endpoints \mbox{rather than within-update trajectories.}

Figure~\ref{fig:overview} additionally compares Qwen3-0.6B and Qwen3-1.7B
under the same MedQA setting, using 1,024 incoming examples and five-seed
means for SFT, Replay, and Task-Bank Self-Replay. Incoming-task gains include
standard errors at all three sizes, while prior-task standard errors are available
for the 4B scores.

Capability changes are computed against each update block's fixed base row.
For capability $j$, we report $100(\bar a_{j,\mathrm{post}}-a_{j,\mathrm{base}})$
in percentage points, with the post-score SE rescaled by 100.
The held-out mean averages the four held-out capability-mean changes, and the held-out
worst takes their minimum. This minimum of means differs from the mean of
per-seed minima in Table~\ref{tab:protection}. The latter cannot be recovered
from aggregate scores. We therefore report no derived SE for the
mean or worst profile, no paired significance test, and no forgetting among
initially \mbox{correct examples for this study.}

\begin{table}[!ht]
\centering\small\setlength{\tabcolsep}{3.5pt}
\caption{\textbf{Broad replay can limit prior-capability losses while maintaining new-task gains.} Single-Task Replay uses one prior target. Replay and Task-Bank Self-Replay use the four-task inventory: Replay uses stored examples and labels, and Task-Bank Self-Replay uses model-generated questions and answers. Qwen3-4B, learning rate $\learningrate{1e-5}$, one epoch, five seeds. Scores are absolute accuracy percentages. Pre-update scores come from the initial checkpoint. Other entries are endpoint means $\pm$ SE.}
\label{tab:manual_cstance}
\begin{tabular}{lrrrrr}
\toprule
Task & Pre-update & SFT & \shortstack{Single-Task\\Replay} & Replay & \shortstack{Task-Bank\\Self-Replay} \\
\midrule
\multicolumn{6}{l}{\emph{New task}} \\
C-STANCE & $52.44$ & $80.21\pm0.44$ & $80.18\pm0.72$ & $80.30\pm0.83$ & $80.29\pm0.75$ \\
\addlinespace
\multicolumn{6}{l}{\emph{Displayed replay tasks}} \\
MMLU & $64.26$ & $63.94\pm0.68$ & $64.92\pm0.50$ & $64.85\pm0.20$ & $64.89\pm0.30$ \\
HellaSwag & $73.92$ & $67.30\pm0.69$ & $69.01\pm0.53$ & $74.30\pm0.20$ & $73.50\pm0.40$ \\
BoolQ & $85.10$ & $79.00\pm0.20$ & $80.20\pm1.00$ & $84.90\pm0.00$ & $85.10\pm0.10$ \\
\addlinespace
\multicolumn{6}{l}{\emph{Held-out evaluation tasks}} \\
ARC-Challenge & $87.70$ & $81.60\pm0.59$ & $81.00\pm0.59$ & $88.10\pm0.49$ & $88.00\pm0.00$ \\
WinoGrande & $60.94$ & $55.40\pm0.40$ & $56.90\pm0.10$ & $61.00\pm0.20$ & $61.20\pm0.40$ \\
PIQA & $83.59$ & $79.60\pm0.90$ & $80.10\pm0.70$ & $84.10\pm0.20$ & $84.25\pm0.95$ \\
SVAMP & $32.03$ & $30.30\pm0.39$ & $30.30\pm0.40$ & $34.50\pm0.20$ & $35.10\pm0.20$ \\
\bottomrule
\end{tabular}
\end{table}

\subsection{Confidence and behavioral probes of competence}
\label{app:self_knowledge}

Confidence-based self-assessment provides one possible signal about what a
model can do \citep{kadavath2022know}. Here \emph{competence} means task
accuracy: we compare estimates derived from the model's confidence with
measured accuracy. Task-level \emph{calibration
gap} is the absolute difference between mean confidence and accuracy.
\emph{Expected calibration error} (ECE) summarizes their mismatch across
confidence bins. The area under the receiver operating characteristic curve
(AUROC) measures discrimination between correct and incorrect answers.
Appendix~\ref{app:competence_estimation} gives the confidence and probe protocols.

Better-calibrated answers do not necessarily produce a better ranking of
task-level competence. Qwen3-4B has better item-level discrimination and
calibration than Qwen3-1.7B, yet has lower task-level rank correlation
(Table~\ref{tab:selfknowledge}, panel A). Qwen3.5-4B performs best on the
reported confidence metrics, but its architecture and training generation
also differ. These comparisons do not establish a general advantage for larger models.

Task accuracy can also be estimated from a small set of labeled examples.
Increasing the number of probes reduces accuracy-estimation error on the
disjoint remaining examples for every tested model (panel B).

\begin{table}[ht]
\caption{\textbf{Improved calibration does not guarantee improved task-level competence ranking.} Panel A contrasts task-level estimates with item-level
confidence metrics across seven multiple-choice capabilities. Panel B reports
mean absolute error (MAE) when task accuracy is estimated from $n$ labeled
behavioral probes and tested on the disjoint remainder, over 100 resamples.
Larger probe budgets reduce this error for every model. Arrows indicate the
preferred direction. Calibration gap, ECE, and MAE use the 0--1 accuracy scale,
while Spearman, AUROC, and Brier score are dimensionless. Bold identifies the best
model within each column, not statistical significance.}
\label{tab:selfknowledge}
\centering
\small
\setlength{\tabcolsep}{5pt}
\begin{tabular}{lccccc}
\toprule
\multicolumn{6}{l}{\textbf{A. Confidence-based competence estimates}} \\
\cmidrule(lr){2-3}\cmidrule(lr){4-6}
 & \multicolumn{2}{c}{Task-level} & \multicolumn{3}{c}{Item-level} \\
Model & Spearman $\uparrow$ & Gap $\downarrow$ & AUROC $\uparrow$ & ECE $\downarrow$ & Brier $\downarrow$ \\
\midrule
Qwen3-1.7B   & 0.679 & 0.328 & 0.681 & 0.331 & 0.332 \\
Qwen3-4B     & 0.607 & 0.235 & 0.756 & 0.238 & 0.239 \\
Qwen3.5-4B   & \textbf{0.857} & \textbf{0.059} & \textbf{0.787} & \textbf{0.069} & \textbf{0.139} \\
\bottomrule
\end{tabular}

\medskip
\begin{tabular}{lcccc}
\toprule
\multicolumn{5}{l}{\textbf{B. Accuracy-estimation MAE from labeled probes} $\downarrow$} \\
Model & $n{=}1$ & $n{=}4$ & $n{=}16$ & $n{=}64$ \\
\midrule
Qwen3-1.7B   & 0.399 & 0.176 & 0.089 & 0.046 \\
Qwen3-4B     & 0.343 & 0.163 & 0.079 & 0.043 \\
Qwen3.5-4B   & \textbf{0.263} & \textbf{0.137} & \textbf{0.071} & \textbf{0.037} \\
\bottomrule
\end{tabular}
\end{table}

\subsection{Supplied-question Task-Bank Self-Replay protocol}
\label{app:self_replay}

In the fixed-example-allocation study, the frozen reference model constructs
replay answers from supplied questions and verified demonstrations. For each replay task, 64 replay prompts are
sampled deterministically. Each sampled question is paired with two distinct
within-task demonstrations formatted as a prompt and verified response. The
reference model receives those examples followed by the sampled question.
For multiple-choice tasks it selects the minimum-loss candidate, and for
numeric tasks it generates at most 64 tokens. The generated answer is used for replay. The sampled question's source label is never used to choose or filter
that answer, while the demonstration responses condition the model on the task.

Generation checks enforce valid formatting and access boundaries, not answer
correctness. Empty generations, out-of-set multiple-choice answers, excessive
sequence length, or insufficient distinct demonstrations fail closed. The
generated file records demonstration identifiers, source and output hashes,
the model revision, and a declaration that held-out tasks were not exposed.

Archived manifests connect the generated replay data to the reported training
runs. The generation manifest covers 256 examples across four replay tasks
and has SHA-256 prefix \texttt{ade2db486584}. The two completed training
cells have prefixes \texttt{b437e435e445} and \texttt{af72b9d18a89}. Full
digests are recorded in the release manifest. Both use the same
197-incoming/59-replay row split as Replay.

\section{Additional Related Work}
\label{app:related}

\paragraph{Continual learning with structured interfaces.}
Continual learning can use explicit concepts to preserve interpretable
predictions during adaptation. Tree of Concepts uses a shallow decision tree
to define a fixed concept interface and updates the concept extractor and
prediction head under clinical distribution shift
\citep{cho2026treeconceptsinterpretablecontinual}. Here, Task Banks provide
an interface for specifying replay targets, and we evaluate whether their
protection extends to held-out behaviors.

\paragraph{Unsupervised and task-free continual learning.}
An incomplete preservation inventory is compatible with fully supervised
incoming-task learning. Self-supervised continual learning learns
representations from sequences of unlabeled inputs \citep{yang2023sscl}.
Task-free and task-agnostic methods remove explicit task definitions or
boundaries from the incoming stream
\citep{wang2022taskfree,ardywibowo2022varigrow}. Our incoming data remain
labeled and their task boundary known. The limited information concerns
prior capabilities and differs by replay interface. Replay uses stored
examples and labels. Task-Bank Self-Replay generates answers to supplied
questions in the fixed-allocation study, and generates both questions and
answers in the endpoint, \mbox{bank-composition, and target-selection studies.}

\paragraph{Self-generated and on-policy supervision.}
Self-generated data can supply learning signals without a separately labeled
answer for every generated example. Self-Instruct generates instructions,
inputs, and outputs, then filters invalid or redundant examples before
fine-tuning \citep{wang2023selfinstruct}. LAMOL uses generated pseudo-examples
to rehearse earlier tasks \citep{sun2019lamol}. Even filtered model-generated examples can contain errors
\citep{wang2023selfinstruct}. Their value for replay depends on whether they preserve earlier
behavior while reducing dependence on stored examples \citep{marek2026selfreplay}.

Self-generated supervision also supports adaptation through model-derived
teachers and learning signals. Language models bootstrap from
high-confidence solutions on unlabeled prompts \citep{huang2022selfimprove},
and self-play fine-tuning contrasts model-generated responses with
demonstrations \citep{chen2024spin}. Related methods use
demonstration-conditioned or current-policy teachers
\citep{shenfeld2026sdft,zhao2026opsd,chen2026onpolicy}, model-generated update
directives \citep{zweiger2025seal}, or synthetic ``dreaming'' curricula
\citep{behrouz2026sleep}.

\looseness=-1
We use task descriptions and examples to guide replay generation, then
study supplied-bank composition and target selection for an incoming update.
Held-out evaluation tests whether this replay benefits behavior outside the
supplied targets. Our exploratory bank-expansion pilot separately examines model-proposed
additions.

\looseness=-1
\paragraph{Task generation, capability discovery, and preservation breadth.}
Research on model-generated tasks extends beyond replay within a single
supplied target. Self-Instruct generates diverse instructions and their
input--output examples for instruction tuning \citep{wang2023selfinstruct}.
Automated Capability Discovery uses a scientist model to propose tasks and
probe a subject model, including itself. Its objective is to discover and
measure capabilities rather than protect them during an update
\citep{lu2025acd}. Absolute Zero learns a task-proposal and solution curriculum
with code-execution rewards to improve reasoning \citep{zhao2025absolutezero}.
SEAL generates data and update directives for adaptation
\citep{zweiger2025seal}. These directions motivate model-generated inventories
and learning signals, but task diversity or discovery alone does not establish
preservation of \mbox{prior correct behavior after fine-tuning.}

\looseness=-1
Prior preservation methods can also benefit tasks beyond their supplied
examples. CPLUS uses self-probe gradients to limit interference and evaluates
protection on benchmarks not represented in training
\citep{cho2026continuallearningselfprobegradients}. Our focus is how a partial preservation
inventory can organize model-generated evidence to protect a broader range
of prior behavior. We study replay across supplied targets, combinations of
banks, and selection of entries for an incoming update. The bank-expansion
pilot provides an initial investigation of model-proposed additions, but
establishing their preservation value relative to additional replay within
the existing inventory remains an open question.

\looseness=-1
\paragraph{Capability self-knowledge.}
Evidence about competence and update interference can inform self-preservation
without implying
privileged introspection. Language models can
sometimes predict whether their own answers are correct, although calibration
weakens under distribution shift \citep{kadavath2022know}. Our auxiliary
diagnostics compare confidence-based competence estimates with measured task
accuracy. Preservation is evaluated separately after the update. Estimating
present competence and predicting which capabilities an update will \mbox{harm are distinct questions.}

\section{Task Bank Interfaces and Auxiliary Diagnostics}
\label{app:bank_details}

\begin{table}[!htbp]
\centering
\small
\setlength{\tabcolsep}{4pt}
\begin{tabular}{p{0.24\linewidth}p{0.31\linewidth}p{0.37\linewidth}}
\toprule
Replay interface & Supplied to the model & Model contributes \\
\midrule
Replay & Replay questions and labels & No generated replay answer \\
Supplied-question answers & Demonstrations and unlabeled questions & Answers to supplied questions \\
Question-and-answer generation & Supplied task entries & Replay questions and answers \\
Exploratory expanded bank & Incomplete task inventory & Additional task entries, then questions and answers \\
\bottomrule
\end{tabular}
\caption{\textbf{Replay designs differ in the preservation information they require.}
These are alternative interfaces. The endpoint and supplied-bank
composition/selection studies generate both questions and answers. The
fixed-example-allocation study instead generates answers to supplied
questions. Model-proposed additions are examined in the \mbox{separate exploratory bank-expansion pilot.}}
\label{tab:bank_variants}
\end{table}

\subsection{Prior capabilities, Task Banks, and held-out evaluation}

\looseness=-1
We formalize the prior capabilities that self-preservation aims to retain.
Let $\mathcal{R}$ denote the subset of the model's pre-update capabilities
whose performance should survive the update. A common continual-learning
setting exposes $\mathcal{R}$ through named tasks or representative data.
Our setting assumes that no exhaustive enumeration is available.

The Task Bank represents part of $\mathcal{R}$ through task descriptions and
examples, without claiming to enumerate it. We represent bank entry $j$ by
$b_j=(d_j,E_j,U_j)$: a textual description $d_j$, a small demonstration set
$E_j$, and an optional source of unlabeled prompts $U_j$. A \emph{fixed bank}
$\bank=\{b_1,\ldots,b_m\}$ defines the accessible replay inventory. An entry
may support overlapping skills rather than certify a distinct capability.

The frozen reference model supplies replay through two generation interfaces.
It either answers supplied prompts conditioned on demonstrations, or
generates both questions and answers from task descriptions. These interfaces differ in how many questions, answers, and task
descriptions are externally supplied (Table~\ref{tab:bank_variants}).

A held-out evaluation tests whether the bank supports preservation beyond its own
entries. Let $\audit=\{\mathcal{C}_{m+1},\ldots,\mathcal{C}_{m+r}\}$ denote
the evaluation tasks excluded from bank construction and replay selection.
A method may not use held-out examples, reference answers, task descriptions, or
scores to steer these decisions. This access boundary is distinct from the
study-specific hyperparameter procedures in Appendix~\ref{app:hyperparameters}.
An independently proposed task may
overlap a capability measured by the held-out evaluation without receiving its data.
Performance on $\audit$ tests whether bank-generated replay reduces forgetting. It does
not by itself establish a complete inventory of the model's capabilities.

\subsection{Estimating competence from pre-update predictions}
\label{app:competence_estimation}

As an auxiliary diagnostic, we test whether likelihoods from the pre-update model reveal its current
competence. The seven multiple-choice tasks are ARC-C, BoolQ, C-STANCE,
HellaSwag, MMLU, PIQA, and WinoGrande. These diagnostics are not used to
select replay targets. For a multiple-choice item with candidate sequence losses
$\ell_{i1},\ldots,\ell_{iq}$, choice confidence is
\begin{equation}
  c_i = \max_k \frac{\exp(-\ell_{ik})}
  {\sum_{k'=1}^{q}\exp(-\ell_{ik'})}.
\end{equation}
We compare these confidence scores with observed correctness.
At the item level, AUROC measures discrimination between
correct and incorrect answers, Brier score measures probabilistic prediction
error, and expected calibration error (ECE) measures calibration.
At the task level, Spearman correlation and absolute
calibration gap compare mean confidence with measured accuracy.

A limited probe budget gives an approximate estimate of task competence.
For budget $n$, we sample $n$ labeled examples from a capability and
estimate accuracy by the mean correctness of the model's predictions on
those examples. We measure absolute error
against accuracy on the disjoint remainder and repeat the split 100 times
to estimate error at each probe budget.

\subsection{Replay mixture and fixed-example allocation}
\label{app:replay_training}

The training mixture allocates examples between incoming data and selected
replay entries. Given selected entry indices $S\subseteq\{1,\ldots,m\}$, it is
\begin{equation}
  \mathcal{D}_{\mathrm{mix}} =
  (1-\rho)\,\update + \rho\!\sum_{j\in S}w_j\mathcal{R}_j,
  \qquad \sum_{j\in S}w_j=1,
\end{equation}
where $\update$ and $\mathcal{R}_j$ are the empirical distributions of incoming
examples and replay for entry $j$, $\rho$ is the nominal replay fraction,
and $w_j$ is each selected entry's share of replay. The learner minimizes
$\mathcal{L}_{\mathrm{SFT}}(\theta;\mathcal{D}_{\mathrm{mix}})$ with respect
to its parameters $\theta$, while the reference model remains frozen.
These weights describe example allocation. Tokenization, supervised token
positions, and loss normalization determine the contribution to the training
loss. Allocation and optimization settings are study-specific.

At fixed nominal input allocation, replay replaces some new-task examples.
The fixed-allocation study uses $\rho=59/256$ before tokenization
(59 replay rows and 197 new-task rows), while keeping the optimizer-step
count fixed. The reported comparison includes
SFT, random single-task replay, Replay, Task-Bank Self-Replay, and the
post-hoc selector. Replay arms share
the same nominal row allocation, initialization, learning rate, seeds, and
update count. Token exposure and replay-generation overhead are not matched.

The post-hoc selector provides an analysis-only reference for sparse replay.
It ranks replay tasks by realized SFT damage $-\Delta_j$ and selects the most
damaged task. Its access to post-update damage distinguishes it from a
selection rule available before training. It does not guarantee maximal replay benefit.

Task-Bank Self-Replay changes the replay answers while keeping this
allocation fixed. It uses the demonstration-conditioned procedure in
Appendix~\ref{app:self_replay}, with generation completed before training and
held-out information excluded. This comparison tests generated answers to
supplied questions, not newly invented \mbox{questions or task entries.}

\subsection{Criteria for interpreting target discovery}
\label{app:coverage_validation}

Useful proposals require distinct behavior, demonstrated competence,
and added preservation value.

\paragraph{Distinct behavior.}
A different name or description does not certify new coverage. Structural
validity and exact duplicate filtering do not detect all semantic overlap,
including overlap between separate proposals. A coverage audit must compare
the proposed behaviors with both the initial inventory and the other
additions, using independent behavioral probes and criteria fixed before
post-update evaluation. Model explanations are candidate rationales,
\mbox{and do not certify novelty.}

\paragraph{Pre-update competence and faithful replay.}
The frozen model must demonstrate the proposed behavior on independent
examples. A self-reported ability or calibrated confidence is insufficient.
Generated instances must elicit the behavior claimed by the entry: recognizing
a correct code fragment is not equivalent to writing code that passes tests.
The current four-option generator supplies a restricted recognition format.
Without behavior-specific validation, its results cannot certify preservation
of arbitrary generative or interactive skills.

\paragraph{Incremental preservation value.}
Distinct, answerable tasks can still provide redundant or inefficient replay.
A causal interpretation requires the same adapted model and replay generator,
with bank-aware additions compared against equal-budget replay generation
within the partial bank and unguided additions. Incoming examples and order, update counts,
loss weighting, replay tokens, and seeds must be controlled or their
differences disclosed. Evaluation tasks remain sealed from proposal selection
and generation. The relevant outcomes are Retention (correct $\to$ correct), incoming-task
gain, per-task changes, and paired uncertainty. A single operating point does not
establish an acquisition--retention frontier. Comparable protection with less
learning is not evidence of better target selection.

\subsection{Negative bank-aware expansion pilot}
\label{app:bankaware_negative_pilot}

The bank-aware expansion pilot loses accuracy on the incoming task and all
four held-out tasks. This Qwen3-1.7B seed-1 pilot adapts the model to MedQA with 1,024
incoming examples and 256 generated replay examples, learning rate
$\learningrate{1e-5}$, and 80 optimizer updates. It uses the bank-aware
expanded inventory with an availability-balanced development replay pool.
The final saved model and the unchanged pre-update model are
evaluated on identical 512-example sets per endpoint. Table~\ref{tab:bankaware_negative}
reports matched before/after outcomes. This pilot is analyzed \mbox{separately from the other protocols.}

\begin{table}[!htbp]
\centering
\small
\setlength{\tabcolsep}{6pt}
\begin{tabular}{lrrrr}
\toprule
Task & Base (\%) & Post (\%) & Change (pp) & Retention (\%) \\
\midrule
MedQA test & 43.95 & 39.45 & $-4.49$ & 47.11 [40.44, 53.34] \\
ARC-C & 74.80 & 67.77 & $-7.03$ & 78.59 [74.41, 82.51] \\
WinoGrande & 54.30 & 51.37 & $-2.93$ & 49.28 [43.17, 55.76] \\
PIQA & 64.65 & 61.72 & $-2.93$ & 45.62 [40.79, 51.06] \\
SVAMP & 78.32 & 19.53 & $-58.79$ & 22.69 [18.95, 26.93] \\
\bottomrule
\end{tabular}
\caption{\textbf{The expanded-bank pilot loses new-task accuracy and held-out behavior at the tested operating point.}
The pilot uses Qwen3-1.7B on MedQA with seed 1 and 512 matched examples per endpoint.
Retention (correct $\to$ correct) conditions on initially correct answers.
Brackets are 95\% paired-example bootstrap intervals (1,000 resamples),
which describe evaluation uncertainty within this run. Accuracy changes are
post minus base. Mean held-out change is $-17.92$ pp. This single-arm result
does not measure expansion's effect relative to a fixed bank.}
\label{tab:bankaware_negative}
\end{table}

The negative outcome includes output-format effects. The unchanged SVAMP
numeric extractor obtains no valid final number on 92 base responses and 303 post-update
responses, out of 512. These remain scored as incorrect. We retain this
failure and do not replace the extractor after inspecting the result.
These extraction failures prevent interpreting the accuracy loss solely as
loss of latent mathematical competence. Lower teacher-forced answer loss does not establish
preservation of generated behavior.

Actual supervised exposure is 10,074 incoming tokens
and 2,946 replay tokens, totaling 13,020. This single-arm outcome describes
the tested operating point rather than the relative value of adding targets.

\subsection{Prompt templates for replay, target selection, and exploration}
\label{app:task_bank_prompts}

The templates below describe three interfaces: generating replay, selecting
entries from a supplied bank, and proposing candidate additions. These
operations occur before fine-tuning. Selection returns existing entry IDs,
and exploration returns candidate task descriptions. Braced fields denote the
inputs supplied to each interface. Line wrapping is typographic. We use
simple prompts to test the specified model behaviors. We do not claim that these templates are optimal. The experiments do not
test robustness to alternative prompt wording. The selection templates state the information
available to the selector and the required output.

\paragraph{Selection within a supplied inventory.}
\label{app:target_selection}
Target selection chooses existing entries before replay generation.
Figure~\ref{fig:bank_selection_heldout} compares selection from the four-entry
inventory \{MMLU, GSM8K, HellaSwag, BoolQ\}. Each sparse condition uses two
entries. The full-bank reference uses all four, and the random condition samples
a pair uniformly. These choices do not add entries to the inventory.

The bank-only interface supplies entry IDs, names, and descriptions:

\begin{Verbatim}[fontsize=\small,breaklines=true,breaksymbolleft={},frame=single,framesep=2mm,samepage=true]
Choose exactly two preservation targets from the supplied Task Bank.
Choose targets whose replay could protect a broad range of useful prior capabilities during a future update. Explain briefly why each target is useful.
Return the two selected task IDs and your reasons.

Task Bank: {bank entries}
\end{Verbatim}

Incoming-aware selection uses the same bank inventory. Its template also
supplies an incoming-task description and a fixed sample of training examples:

\begin{Verbatim}[fontsize=\small,breaklines=true,breaksymbolleft={},frame=single,framesep=2mm,samepage=true]
We will fine-tune this model on the incoming task described below.

Choose exactly two preservation targets from the supplied Task Bank.
Choose targets whose replay could protect useful prior capabilities during this update. Explain briefly why each target is useful.
Return the two selected task IDs and your reasons.

Task Bank: {bank entries}
Incoming task: {description}
Training examples: {fixed sample}
\end{Verbatim}

For the reported selection comparison, the incoming task is MedQA. Held-out
evaluation examples, labels, descriptions, and scores are excluded. The
selected IDs define $S$ before replay generation. The frozen reference model
then generates questions and answers for those entries. All conditions use
the Figure~\ref{fig:overview} default training and replay-allocation setting.
These templates specify the selection interface rather than a predetermined
pair of tasks. The choice of two targets does not by itself imply matched
replay tokens or reduced training cost.

\paragraph{Depth: answers to supplied questions.}
The demonstration-conditioned protocol in Appendix~\ref{app:self_replay}
uses two within-task demonstrations and a supplied target question:

\begin{Verbatim}[fontsize=\small,breaklines=true,breaksymbolleft={},frame=single,framesep=2mm,samepage=true]
Examples of the capability to preserve:

Example 1
{demonstration prompt 1}
Verified response: {demonstration response 1}

Example 2
{demonstration prompt 2}
Verified response: {demonstration response 2}

Now answer the new instance in the same way.
{target prompt}
\end{Verbatim}

\looseness=-1
The bank selects the task and its demonstration pool. This template receives
examples rather than a separate task name or description. Only the
demonstration responses are supplied as verified answers. The target answer
is withheld. The generated answer is then used for replay. Thus, this
protocol generates \mbox{answers to existing questions through ICL.}

\paragraph{Depth: synthetic questions and answers.}
For description-based generation, the model receives the entry's description
and generates both a question and an answer:

\begin{Verbatim}[fontsize=\small,breaklines=true,breaksymbolleft={},frame=single,framesep=2mm,samepage=true]
Demonstrate this capability: {description}
Create a new self-contained multiple-choice question and its answer. There must be four nonempty distinct options and one answer letter A, B, C, or D. Return only JSON with fields question (string), options (four strings), answer (letter). Do not claim that the answer has been externally verified. Create a different instance for variation {seed}, attempt {attempt}.
\end{Verbatim}

Here, \texttt{description} specifies the task's scope. The \texttt{seed} and
\texttt{attempt} fields distinguish generation requests. This template uses
the description alone, without in-context demonstrations or a separate
task-name field. The output supplies a synthetic input--output pair.
Formatting checks require a nonempty question, four distinct nonempty
options, and a valid answer letter. Answers are not externally verified.

\paragraph{Exploration: unguided task proposals.}
The model proposes a candidate entry without seeing a bank:

\begin{Verbatim}[fontsize=\small,breaklines=true,breaksymbolleft={},frame=single,framesep=2mm,samepage=true]
Propose one specific capability that you can demonstrate with a question and answer. Explore a diverse capability of your own. No existing capability inventory is supplied.
Return only JSON with string fields name and description. Use a concrete skill rather than a benchmark name. Exploration variant {seed}, attempt {attempt}.
\end{Verbatim}

\paragraph{Exploration: bank-aware task proposals.}
The model sees the retained inventory and proposes additions:

\begin{Verbatim}[fontsize=\small,breaklines=true,breaksymbolleft={},frame=single,framesep=2mm,samepage=true]
Propose one specific capability that you can demonstrate with a question and answer. Inspect this preservation bank and propose useful coverage it does not already represent:
{current bank JSON}
Return only JSON with string fields name and description. Use a concrete skill rather than a benchmark name. Exploration variant {seed}, attempt {attempt}.
\end{Verbatim}

The current-bank field is a JSON list containing only each retained entry's
\texttt{id} and \texttt{description}. Omitted entries, held-out evaluation
inputs and labels, and evaluation scores are excluded. Each bank-aware
request sees the same initial retained bank. Earlier proposals are not
inserted into subsequent prompts.

\paragraph{From proposals to replay.}
Both exploration templates return a name and description. Example pairs
are generated afterward by applying the synthetic question-and-answer
template to each selected description. Proposal filtering checks structural validity and
removes exact duplicate name--description pairs within the proposal stream.
It does not establish semantic novelty or competence. Consequently, an entry
is a \emph{candidate preservation target}. Testing its value requires measuring how much preservation its replay
adds, with the controls in Appendix~\ref{app:coverage_validation}. Proposals need not recover the names
of deliberately omitted tasks.

\subsection{Complete results at fixed training allocation}

The complete comparison reports accuracy changes and forgetting among
initially correct answers alongside new-task gain under the same nominal
allocation (Table~\ref{tab:protection}). Its worst-change columns
average the minimum within each seed, rather than taking the minimum of
capability means as in the separate five-seed endpoint study. This distinction
matters when the identity of the worst-affected capability varies between runs.

\begin{table}[!ht]
\caption{\textbf{Generated replay reduces measured forgetting alongside smaller new-task gains at a fixed example allocation.} Qwen3-4B is adapted to
MedQA at $\learningrate{1e-6}$ for one epoch with two seeds per arm. All arms allocate
256 input rows. Replay occupies 59 rows before tokenization, replacing new-task
inputs. Worst is the mean of each seed's minimum capability change within the
indicated task group. Retention (correct $\rightarrow$ correct) equals
$100-\mathrm{MF}$ on the same eligible tasks. MF is mean forgetting among initially correct examples over tasks
with enough initially correct examples under the rule in Appendix~\ref{app:measurements}.
It need not cover every task included in Worst. Higher changes and lower MF
are better. The post-hoc selector is analysis-only. Bold identifies the best
reported retention summary, not statistical significance.}
\label{tab:protection}
\centering
\small
\setlength{\tabcolsep}{4pt}
\begin{tabular}{lrrrrrr}
\toprule
 & & & \multicolumn{2}{c}{Replay tasks} & \multicolumn{2}{c}{Held-out evaluation} \\
\cmidrule(lr){4-5}\cmidrule(lr){6-7}
Method & New-task & Replay & Worst & MF & Worst & MF \\
 & gain (pp) & rows & (pp) & (\%) & (pp) & (\%) \\
\midrule
SFT & +1.17 & 0 & -3.04 & 3.68 & -1.56 & 5.74 \\
Random replay & +0.98 & 59 & -1.37 & 3.02 & -1.76 & 5.00 \\
Task-Bank Self-Replay & +0.49 & 59 & \textbf{-0.39} & \textbf{2.03} & \textbf{-0.68} & \textbf{3.52} \\
Replay & +0.98 & 59 & -1.27 & 2.26 & -0.88 & 3.82 \\
Post-hoc selector & +1.08 & 59 & -0.59 & 2.42 & -1.27 & 5.01 \\
\bottomrule
\end{tabular}
\end{table}
\FloatBarrier
\section{Full Reported Learning-Rate Sweeps}
\label{app:lr_profiles}

\begin{figure}[!htbp]
\centering
\includegraphics[width=\linewidth]{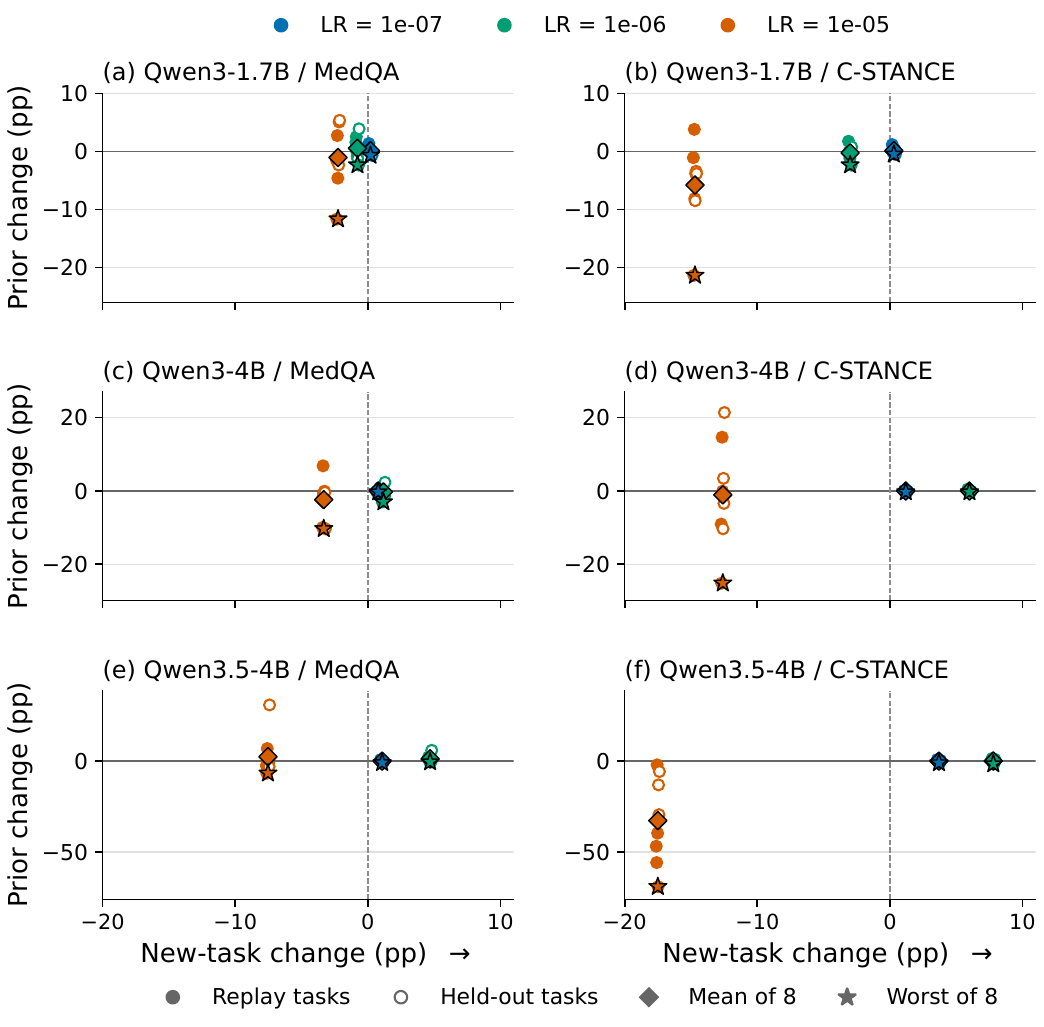}
\caption{\textbf{An update can damage prior capabilities without improving performance on the new task.} Rows identify models and columns identify update tasks.
Color identifies learning rate. Rightward is greater new-task gain, and upward
is a more positive prior-task accuracy change. Each circle is one capability's change averaged across
two seeds: filled circles denote bank tasks and open circles
denote held-out tasks. Diamonds and stars summarize the mean and minimum of
these eight capability means. Circles have small horizontal offsets for
visibility. Summary markers retain the exact new-task gain coordinate. Zero
denotes no change from the pre-update model. Vertical scales match within model rows but
differ across rows. All main sweep learning-rate conditions are retained.}
\label{fig:frontier}
\end{figure}

The controlled sweep separates negligible adaptation, useful new-task gain,
and damage to both new and prior tasks (Figure~\ref{fig:frontier}). Retaining
every measured learning rate makes clear why small prior-task accuracy changes must be
interpreted alongside new-task gain. The destructive conditions are stress tests,
not the sole explanation for prior-capability losses during useful learning.

\begin{figure}[!htbp]
\centering
\includegraphics[width=\linewidth]{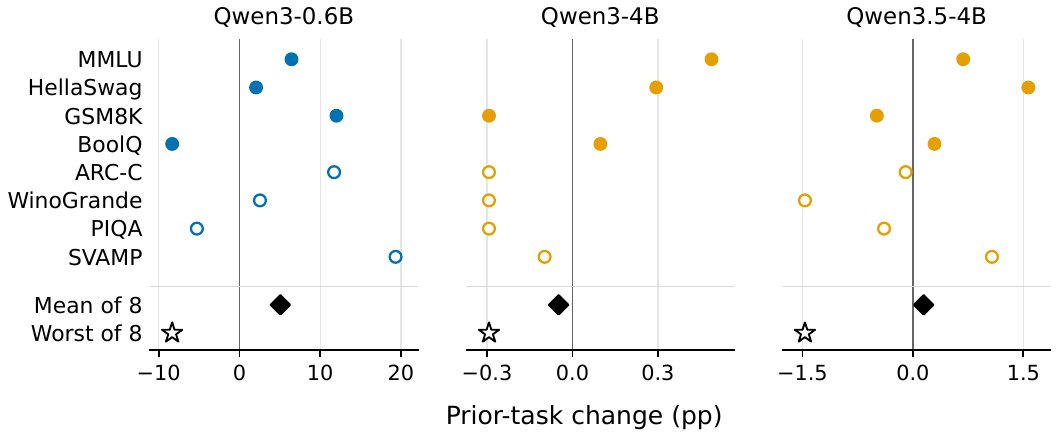}
\caption{\textbf{Positive average changes can conceal losses on individual prior tasks even when the incoming task improves.} Qwen models after C-STANCE SFT at selected operating
points with new-task gains of $+7.1$, $+6.0$, and $+7.8\pp$, respectively.
Each circle is a task's change averaged across two seeds. Diamonds and stars
show the mean and minimum across eight task means. Filled markers identify
bank tasks and open markers held-out tasks. These labels identify the
evaluation split, not replay in these SFT runs. Blue denotes 0.6B parameters
and orange 4B. Horizontal ranges differ by model, with zero marking the
pre-update score. Compare magnitudes by tick values, not visual distances.
The learning rates are $\learningrate{3e-6}$, $\learningrate{1e-6}$, and $\learningrate{1e-6}$.
The 0.6B panel uses the separate 256-row C-STANCE calibration, not the
1,024-row follow-up. This selected comparison does not isolate the effect of model scale from
model generation and learning rate. The complete grid follows.}
\label{fig:profiles}
\end{figure}

\FloatBarrier

\begin{figure}[!htbp]
\centering
\includegraphics[width=\linewidth,height=0.76\textheight,keepaspectratio]{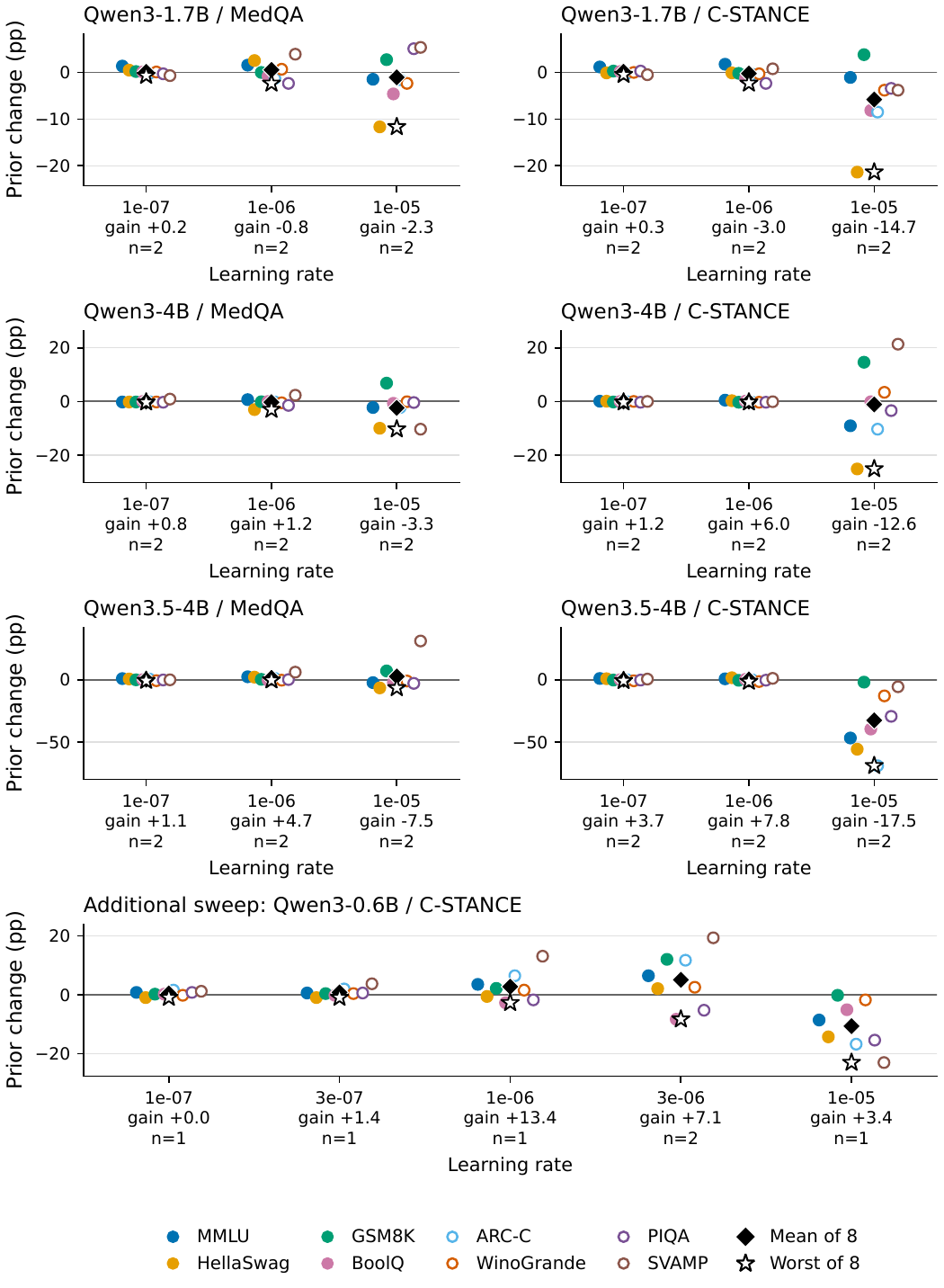}
\caption{\textbf{Learning rate changes both the severity and the distribution of forgetting, with different patterns across models and incoming tasks.} The top three rows show the complete paired main sweep, with model by row and
update task by column. The bottom panel separately shows the additional
Qwen3-0.6B C-STANCE sweep. Circles show capability means, with filled markers for bank tasks
and open markers for held-out tasks. Diamonds and stars are the mean and
minimum of the eight capability means. Labels below each learning rate report
new-task gain in percentage points and seed count $n$. Vertical scales
match within model rows but differ across rows. The additional Qwen3-0.6B
calibration uses 256 rows and covers C-STANCE only. It is separate from the
1,024-row follow-up. Every measured condition is retained, \mbox{including high-rate failures as stress tests.}}
\label{fig:lr_atlas}
\end{figure}

Learning-rate sensitivity depends on the model and incoming task, not just
the average prior-task change. Figure~\ref{fig:lr_atlas} shows each measured
capability profile across the reported SFT conditions, extending the selected
operating points in Figure~\ref{fig:profiles}. The sparse grid describes these
conditions without establishing a scaling law or \mbox{a universal safe learning rate.}

\FloatBarrier
\section{Model Scale and Broad Preservation}
\label{app:scale_preservation}

We summarize the matched MedQA comparison in Figure~\ref{fig:overview}
using the four held-out tasks: ARC-C, WinoGrande, PIQA, and SVAMP.
For task $j$, accuracy loss is $-\Delta_j$ in percentage points.
Average loss is the unweighted mean across these four tasks, and worst-task
loss is their maximum. Negative loss indicates an improvement over the pre-update
model. These summaries are computed from five-seed task means. Worst-task
loss is the worst of those means, rather than a mean of seed-wise maxima.

Replay reduces average and worst held-out losses across the measured sizes,
but the two replay sources show different trends. Average held-out accuracy
loss under SFT decreases
from 6.01 to 4.38 to 3.50 pp at 0.6B, 1.7B,
and 4B, while worst-task loss decreases from 8.30 to 5.94 to 5.40 pp.
Replay reduces both summaries relative to SFT, but retains average losses
of 2.00, 1.27, and 1.25 pp and worst-task losses of 2.81, 1.91, and
1.69 pp at the three sizes. Task-Bank Self-Replay improves the four-task
average by 0.61, 0.31, and 0.76 pp, respectively. Its worst-task accuracy losses
are 0.52, 1.03, and 0.39 pp, so an average accuracy gain does not imply that
every task is preserved. Replay losses decrease across these
measured sizes but remain positive at each size. Task-Bank Self-Replay
does not follow a monotonic scaling pattern here.

\FloatBarrier
\section{Sequential Preservation Pilot}
\label{app:self_improvement}

A two-update pilot separates final gains over the initial model from
accuracy changes on the previously learned task during the second update. Qwen3-0.6B learns MedQA
and then C-STANCE, with seed 0, learning rate $\learningrate{3e-6}$, and one epoch
of 256 rows per stage. Replay arms use 197 new-task and 59 replay rows per
stage. The expanding generated bank adds MedQA before the second update.
The fixed generated bank retains its original inventory.

\begin{table}[!htbp]
\centering
\small
\setlength{\tabcolsep}{4pt}
\begin{tabular}{lrrrrr}
\toprule
 & \multicolumn{4}{c}{Initial $\to$ final change (pp)} & Stage 1 $\to$ 2 (pp) \\
\cmidrule(lr){2-5}\cmidrule(lr){6-6}
Method & MedQA & C-STANCE & Worst replay & Worst held-out & MedQA \\
\midrule
Sequential SFT & +5.68 & +13.00 & -12.55 & -8.79 & +0.39 \\
Replay & +9.39 & +13.60 & -2.35 & +2.54 & +0.59 \\
Task-Bank Self-Replay & +3.91 & +17.20 & -5.49 & -12.70 & +0.20 \\
Task-Bank Self-Replay + MedQA & +3.52 & +16.60 & -9.80 & -3.71 & -0.20 \\
\bottomrule
\end{tabular}
\caption{\textbf{Adding a replay target can improve held-out accuracy while worsening accuracy on replay tasks.}
Adding MedQA to Task-Bank Self-Replay improves worst held-out change
but worsens worst replay-task change relative to the fixed bank.
Qwen3-0.6B learns MedQA then
C-STANCE with one seed. The first four score columns compare the final model
with the initial model. The last compares MedQA accuracy immediately before
and after the C-STANCE update on the same 511 evaluation examples. All changes
are in percentage points. The expanded arm remains above the initial MedQA score but
slightly below its own first-stage score, illustrating why final gains do not
establish retention of every intermediate gain. Stage-relative changes measure
net accuracy, not preservation of individual correct answers. No uncertainty
or significance is inferred from this one-seed pilot.}
\label{tab:sequential_pilot}
\end{table}

The fixed and expanding generated-bank arms share the first-stage checkpoint.
Generation uses 64 candidate prompts per replay task. Replay answers use no
source answer labels or held-out data. Both stage endpoints are recorded for
the comparisons in Table~\ref{tab:sequential_pilot}.

This pilot does not establish continual self-improvement or autonomous
capability discovery: task additions are prescribed and only two stages and
one seed are tested. Longer-run accumulation is a distinct requirement from
the bank-composition and target-selection comparisons in
Sections~\ref{sec:bank_composition} and~\ref{sec:target_selection}.
Agent self-improvement \citep{zhang2025dgm} and meta-learned continual-learning
algorithms \citep{irie2025metalearning} motivate studying longer-run accumulation,
which remains beyond the scope of this pilot.

\FloatBarrier
\section{Further Discussion and Limitations}
\label{app:limitations}

\paragraph{Choosing preservation targets.}
Preservation depends on which targets supply replay and how their examples
are generated. Combining partial banks improves mean held-out Retention,
and incoming-aware selection helps at the tested 4B operating point
(Figures~\ref{fig:bank_width_heldout} and~\ref{fig:bank_selection_heldout}).
Assessing these benefits requires new-task gains, individual-task changes,
and Retention on tasks excluded from replay.

\paragraph{Scope.}
\looseness=-1
The experiments establish preservation behavior only within the reported
models, tasks, and short update sequences. They cover Qwen checkpoints, two
incoming tasks, and eight prior benchmark capabilities, including four
held-out evaluation tasks. Transfer to that held-out evaluation does not establish preservation
of the full pretrained repertoire, and stable low baseline performance can
provide little evidence of retained competence. Most studies concern one
update. The two-stage, one-seed pilot has mixed outcomes and does \mbox{not establish sustained autonomous preservation.}

\paragraph{Controls and cost.}
The experimental controls limit causal and uncertainty claims. Selected
model comparisons vary learning rate and model generation, so they do not
isolate a scaling effect. Two-seed controlled studies and five-seed
endpoint aggregates limit uncertainty analysis. The latter do not support paired
significance tests or within-update trajectories. Under a fixed example
allocation, improved preservation accompanies smaller new-task gains as
replay replaces incoming examples. This study matches nominal input rows
and optimizer steps, while tokenized exposure and generation overhead
differ. These distinctions prevent treating every
cross-study contrast as a \mbox{controlled comparison of a single factor.}

\paragraph{Targets and supervision.}
Task-Bank Self-Replay depends on supplied preservation targets. It reduces
some supervision requirements without eliminating them or certifying
autonomous discovery. Replay can reproduce
model errors, and supplied-question replay still relies on verified
demonstrations. The bank-aware discovery pilot does not establish whether
new proposed targets improve preservation over a fixed partial bank.
The evidence does not establish robustness across independent task omissions,
prompt variants, or alternative task-addition rules.

\end{document}